\documentclass[10pt,twocolumn,letterpaper]{article}

\usepackage{iccv}              
\usepackage{algorithm}
\usepackage{algorithmic}
\usepackage{multirow}
\usepackage{pifont}
\usepackage{amsmath}
\usepackage{amssymb}
\usepackage{xcolor} 
\usepackage{booktabs} 
\usepackage{colortbl} 
\usepackage{array} 
\usepackage{graphicx} 

\definecolor{iccvblue}{rgb}{0.21,0.49,0.74}
\usepackage[pagebackref,breaklinks,colorlinks,allcolors=iccvblue]{hyperref}

\def\paperID{1842} 
\def\confName{ICCV}
\def\confYear{2025}

\title{ATCTrack: Aligning Target-Context Cues with Dynamic Target States \\ for Robust Vision-Language Tracking}

\author{%
\textbf{Xiaokun Feng}$^{1,2}$\thanks{Equal Contribution} \hspace{9pt}
\textbf{Shiyu Hu}$^{3}$\footnotemark[1] \hspace{9pt} 
\textbf{Xuchen Li}$^{1,2,4}$ \hspace{9pt}
\textbf{Dailing Zhang}$^{1,2}$\hspace{9pt}
\\
\textbf{Meiqi Wu}$^{2}$ \hspace{9pt}
\textbf{Jing Zhang}$^{2}$ \hspace{9pt}
\textbf{Xiaotang Chen}$^{1,2}$ \hspace{9pt}
\textbf{Kaiqi Huang}$^{1,2}$ \thanks{Corresponding authors} \hspace{9pt}
\\
$^1$School of Artificial Intelligence, UCAS\\
$^2$The Key Laboratory of Cognition and Decision Intelligence for Complex Systems, CASIA\\
$^3$School of Physical and Mathematical Sciences, NTU  $^4$ZGCA\\\\
}

\begin{document}
\maketitle
\begin{abstract}
Vision-language tracking aims to locate the target object in the video sequence using a template patch and a language description provided in the initial frame.
To achieve robust tracking, especially in complex long-term scenarios that reflect real-world conditions as recently highlighted by MGIT, it is essential not only to characterize the target features but also to utilize the context features related to the target.
However, the visual and textual target-context cues derived from the initial prompts generally align only with the initial target state. 
Due to their dynamic nature, target states are constantly changing, particularly in complex long-term sequences. It is intractable for these cues to continuously guide Vision-Language Trackers (VLTs).
Furthermore, for the text prompts with diverse expressions, our experiments reveal that existing VLTs struggle to discern which words pertain to the target or the context, complicating the utilization of textual cues.
In this work, we present a novel tracker named \textit{\textbf{ATCTrack}}, which can obtain multimodal cues \textbf{A}ligned with the dynamic target states through comprehensive \textbf{T}arget-\textbf{C}ontext feature modeling, thereby achieving robust tracking. 
Specifically,
\textbf{(1)} for the visual modality, we propose an effective temporal visual target-context modeling approach that provides the tracker with timely visual cues.
\textbf{(2)} For the textual modality, we achieve precise target words identification solely based on textual content, and design an innovative context words calibration method to adaptively utilize auxiliary context words.
\textbf{(3)} We conduct extensive experiments on mainstream benchmarks and ATCTrack achieves a new SOTA performance.
The code and models will be released at: \href{https://github.com/XiaokunFeng/ATCTrack}{https://github.com/XiaokunFeng/ATCTrack}.
\end{abstract}    
\section{Introduction}
\label{sec:intro}

\begin{figure}[t!]
    \centering
    \includegraphics[width=\linewidth]{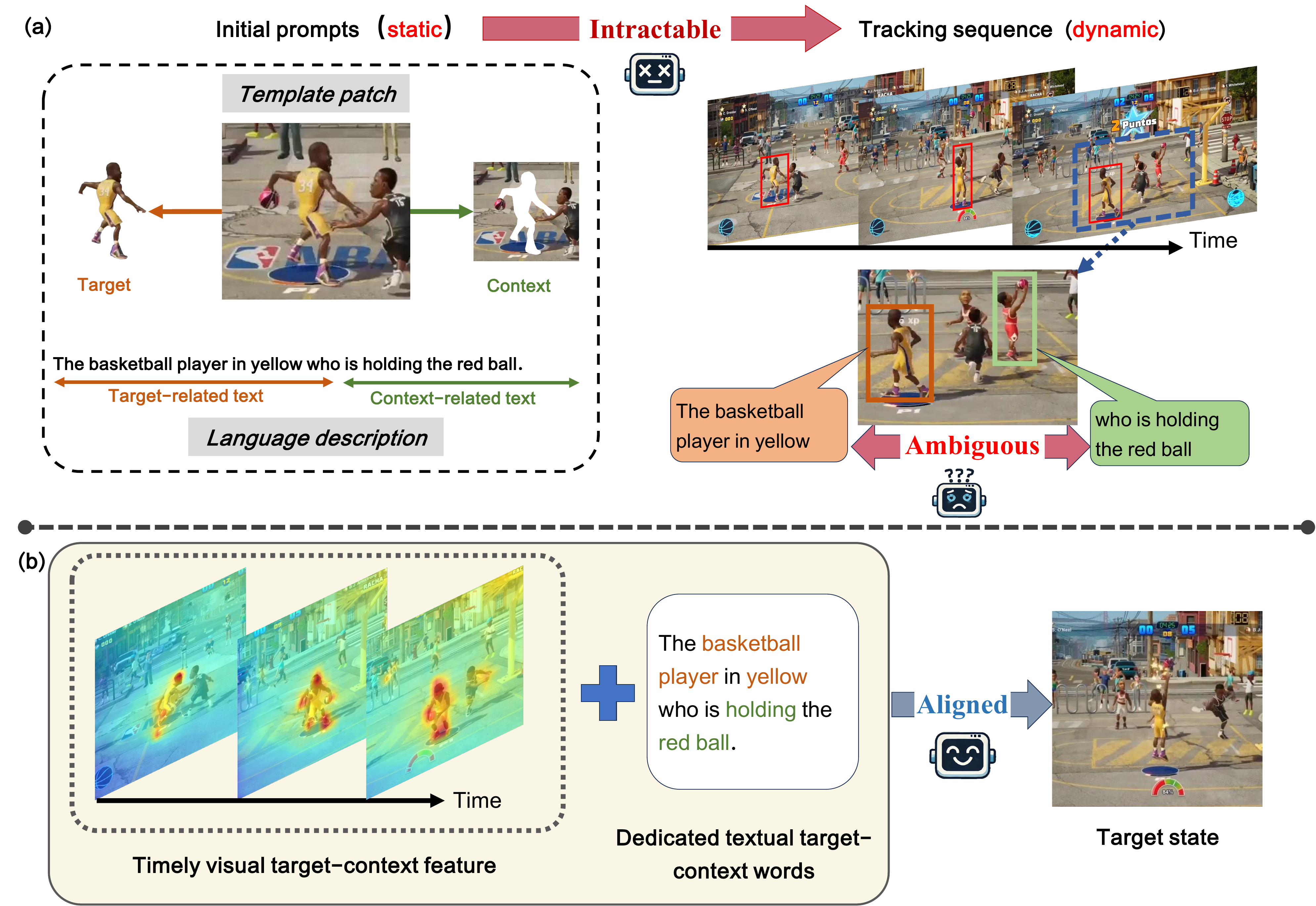}
    \caption{\textbf{(a) Limitations of initially given prompts} (\ie, the template patch and language description).
    Despite containing certain multimodal target-context cues, these static initial prompts are intractable for continuously guiding the tracker in dynamic tracking sequences.
    Particularly, the objects corresponding to the target-related text and context-related text can be ambiguous, which may mislead the tracker.
\textbf{(b) Our key insights lie in providing multimodal target-context cues aligned with the dynamic target states.} For the visual modality, we model timely visual target-context features to adapt to dynamic changes. For the textual modality, we achieve precise awareness of target words and mitigate the potential misleading effects of context words.
    }
    \label{fig1}
\end{figure}

Given a template patch and a language description in the initial frame, Vision-Language Tracking (\textbf{VLT}) task \cite{OTB-Lang} aims to locate a user-defined object in a video sequence. 
Harnessing the complementary advantages of multiple modalities \cite{lahat2015multimodal}, recent studies, exemplified by MGIT \cite{hu2023multi}, seek to explore the performance of Vision-Language Trackers (\textbf{VLTs}) in \textbf{complex long-term} sequences. 
These scenarios, encompassing multifaceted spatiotemporal and causal relationships, more accurately reflect real-world conditions while posing new challenges for tracker design.

To achieve robust tracking within these environments, we first need to utilize the provided visual and textual cues to characterize the \textbf{target feature} \cite{li2022cross}. 
As shown in \cref{fig1} (a), the appearance of the target in the template image and the target-related text (\eg, target class, appearance attributes) in the language description offer basic reference information for tracking.
However, complex long-term scenarios are more commonly accompanied by challenges such as occlusion and similar object distractions \cite{SOTVerse,zhao2024biodrone}, making it insufficient to rely solely on these target cues.
It is also essential to model the \textbf{context feature} of the target \cite{shao2024queryvlt,xu2025less}. Specifically, the visual content surrounding the target and context-related text (\eg, other reference objects) contribute to more robust tracking. 
Therefore, the target and context features collectively depict the \textbf{target states}, and \textbf{\textit{efficiently representing and leveraging target-context features}} is crucial for handling complex long-term scenarios.

Considering that the initially given prompts carry certain multimodal target-context information, most existing VLTs, \eg, VLT$_{\rm{TT}}$ \cite{VLT} and JointNLT \cite{zhou2023joint}, rely entirely on them for tracking. 
For the visual template image cropped from the first frame, its cropping region is determined by enlarging the target's bounding box, thus including some context features \cite{zheng2022leveraging}. For the textual description, the full sentence is utilized as a whole.
Although achieving some effectiveness, they overlook the inherent dynamic nature of the target states. 
In complex long-term scenarios, target states often undergo significant changes.
As shown in \cref{fig1} (a), the subsequent target states gradually deviate from the initially given prompts. 
This misalignment makes the latter \textbf{intractable} for continuously guiding the tracker. 
Furthermore, for the text prompts, the object corresponding to the target-related text and context-related text can be \textbf{ambiguous}, which may significantly mislead the tracker \cite{hu2023multi,li2024dtllm}.

\begin{figure*}[t!]
\vspace{-0.5cm} 
    \centering
    \includegraphics[width=0.9\linewidth]{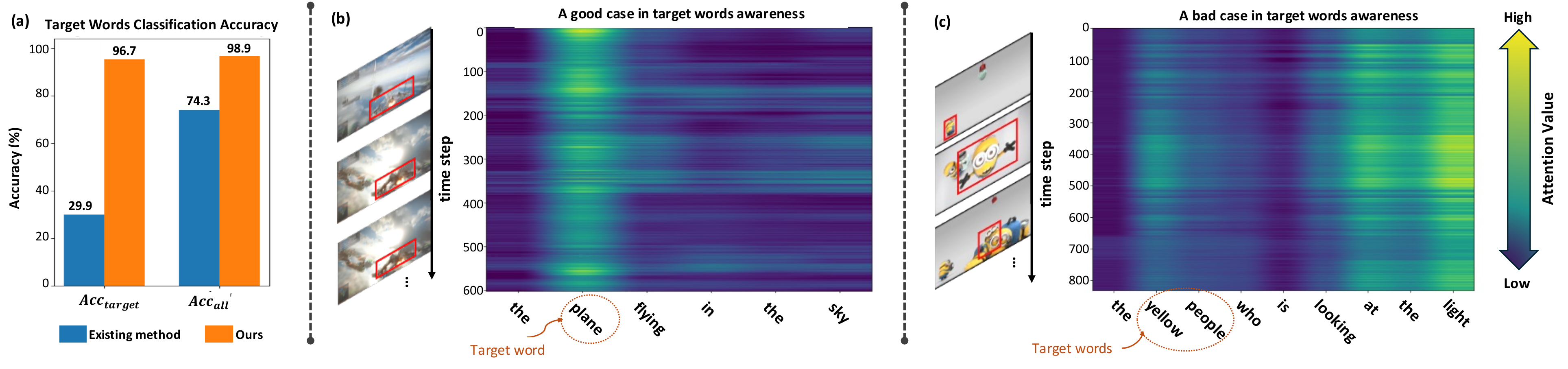}
    \caption{\textbf{(a)} Comparison of the existing vision-text similarity-based method and ours in terms of target words classification accuracy. 
    \textbf{(b-c)} Attention distribution maps for the target words during tracking using the existing method. 
    In case (b), the model focuses on the word that corresponds to the tracking target, \ie, ‘plane,' indicating that the tracker correctly understands the intent embedded in the text prompt, thereby effectively utilizing textual cues. 
    In case (c), the model's focus is on the words ‘the light,’ which does not match the tracking target ‘yellow people,’ potentially leading to misguidance in the tracking process.
    Better viewed in color with zoom-in. 
    }
    \label{fig2}
\end{figure*}

Recently, some VLTs, \eg, QueryNLT \cite{shao2024queryvlt} and  TTCTrack \cite{mao2024textual}, attempt to handle different word components of the text prompts specifically, giving sufficient attention to target words and alleviating interference from context words. 
Although well-motivated, these efforts require trackers to automatically distinguish target and context words through visual-textual feature similarity.
Due to the lack of supervision information and the diverse nature of textual expressions, we find that this method does not yield satisfactory results. 
For an intuitive understanding, \cref{fig2} (b) and (c) illustrate cases of correct and incorrect awareness of the target words using this method (see Appendix B.1 for implementation details). 
Additionally, \cref{fig2} (a) presents the quantitative evaluation (see Appendix B.2) results, showing that the classification accuracy for target words achieved by this method is only 29.9\%.

To address the above issues, we propose a novel tracker named \textit{\textbf{ATCTrack}}, which comprehensively models \textbf{T}arget-\textbf{C}ontext features to obtain multimodal cues \textbf{A}ligned with the dynamic target states, thereby achieving robust tracking.
Our key insights are illustrated in \cref{fig1} (b).
For the visual modality, we design an effective temporal visual target-context modeling approach to provide the tracker with timely visual cues. 
Specifically, we explicitly construct a target-context spatial distribution map and integrate it into the updated temporal memory.

For the textual modality, we first propose a precise  target words awareness method based solely on textual content. Intuitively, even without relying on video data, we can identify target words purely from the text. For example, we can determine that the target words in the text prompt of \cref{fig1} (a) are “basketball player, yellow." 
Compared to existing methods that rely on fine-grained alignment between textual words and visual targets \cite{shao2024queryvlt,mao2024textual}, our approach simplifies the task and achieves an impressive 96.7\% target words classification accuracy (Fig.~\ref{fig2}).
Given the lack of sentence component labels in existing benchmarks \cite{hu2023multi,LaSOT,TNL2K,OTB-Lang}, we design an automated target words annotation pipeline by leveraging the off-the-shelf Large Language Models (\textbf{LLMs}) \cite{touvron2023llama,bai2023qwen} to provide supervisory labels for model training.
Furthermore, leveraging the accurately identified target words, we design an innovative context words calibration mechanism to alleviate the potential misleading impact of context words cues.

Benefiting from these multimodal target-context cues aligned with dynamic target states, ATCTrack significantly outperforms existing SOTAs on mainstream benchmarks (\ie, MGIT \protect\cite{hu2023multi}, TNL2K \protect\cite{TNL2K} and LaSOT\(_{ext}\) \cite{fan2021lasot}). Impressively, ATCTrack-B improves over the existing best results by 6.4\%, 4.3\% and 3.5\% in precision, respectively.
Our contributions are as follows:

\begin{itemize}
\item We propose a novel tracker named ATCTrack, which can obtain multimodal cues aligned with the dynamic target states through comprehensive target-context feature modeling. Compared to the limitations of initial prompts, which typically only align with the initial target state, these aligned cues can guide tracking more robustly.
\item For the visual modality, ATCTrack effectively models temporal visual target-context features to capture timely visual cues. 
For the textual modality, ATCTrack achieves precise awareness of target words and mitigates the potential misleading effects of context words.
\item We conduct extensive experiments on mainstream benchmarks and ATCTrack achieves a new SOTA performance.
\end{itemize}

\section{Related Works}

\label{sec:formatting}

\subsection{Traditional Vision-Language Tracking }
Visual language tracking extends the classic visual single object tracking task by incorporating textual descriptions \cite{OTB-Lang}. 
Many early efforts treat initial prompts as core reference information, utilizing the multimodal target-context cues they contain to guide tracking.
Among them, SNLT \cite{SNLT} employs a general language region proposal network to achieve the interaction between multimodal cues and search features, and then uses an aggregation module to integrate multi-scale feature information. 
To enhance the tracker's ability to align visual and textual modalities, both All-in-One \cite{zhang2023all} and UVLTrack \cite{ma2024unifying} design specialized contrastive losses \cite{tian2020makes} to improve the model's multimodal understanding capability.
While these VLTs achieve some success, they overly rely on static initial cues and overlook the dynamic nature of the target and its context. This poses a challenge for maintaining robust tracking when the target states deviate from the initial cues \cite{feng2025memvlt}.

\subsection{Object-Context Modeling in Tracking}
To accommodate the dynamic changes of target states, many trackers attempt to utilize temporal features \cite{zheng2024odtrack,zheng2025decoupled,zhang2025beyond} to obtain updated target-context cues. 
For the visual modality, GTI \cite{GTI} integrates tracking and grounding tasks, replacing the template with updated grounding results. 
To support longer and denser temporal modeling, some trackers store multiple-step visual temporal features as additional memory. 
A representative approach involves using RoI features \cite{ren2015faster} based on predicted bboxes as memory units, \eg, JointNLT \cite{zhou2023joint} and TrDiMP \cite{wang2021transformer}. 
Given that this localized cropping method can only capture limited visual context information, we represent the global target distribution map and use it to construct temporal visual target-context cues.

In addition to visual modality, recent trackers focus on the unique VLT challenges of misalignment between static textual cues and dynamic target states. 
MemVLT \cite{feng2025memvlt}, inspired by prompt learning \cite{lester2021power,bahng2022visual,zhou2022conditional}, compresses dynamic target features into a small set of tokens and uses them to implicitly modulate static multimodal cues. 
To more explicitly leverage multimodal cues, QueryNLT \cite{shao2024queryvlt} approaches the problem from a target-context perspective,  aiming to obtain accurate cues through mutual modulation between dynamic visual features and textual cues. 
Although QueryNLT shares similar motivations with our work, its key modulation process requires fine-grained alignment between visual targets and textual words in VLTs \cite{zhang2024one}. 
As shown in \cref{fig2} (a), we conduct a quantitative evaluation and find that this method fails to effectively identify target and context words. 
In contrast, we design a accurate target words awareness method directly based on textual content and introduce an effective calibration mechanism to mitigate the misleading effects of context words.

\section{Methodology}
\label{method}
\begin{figure*}[htbp]
\vspace{-0.5cm} 
    \centering
    \includegraphics[width=1\textwidth]{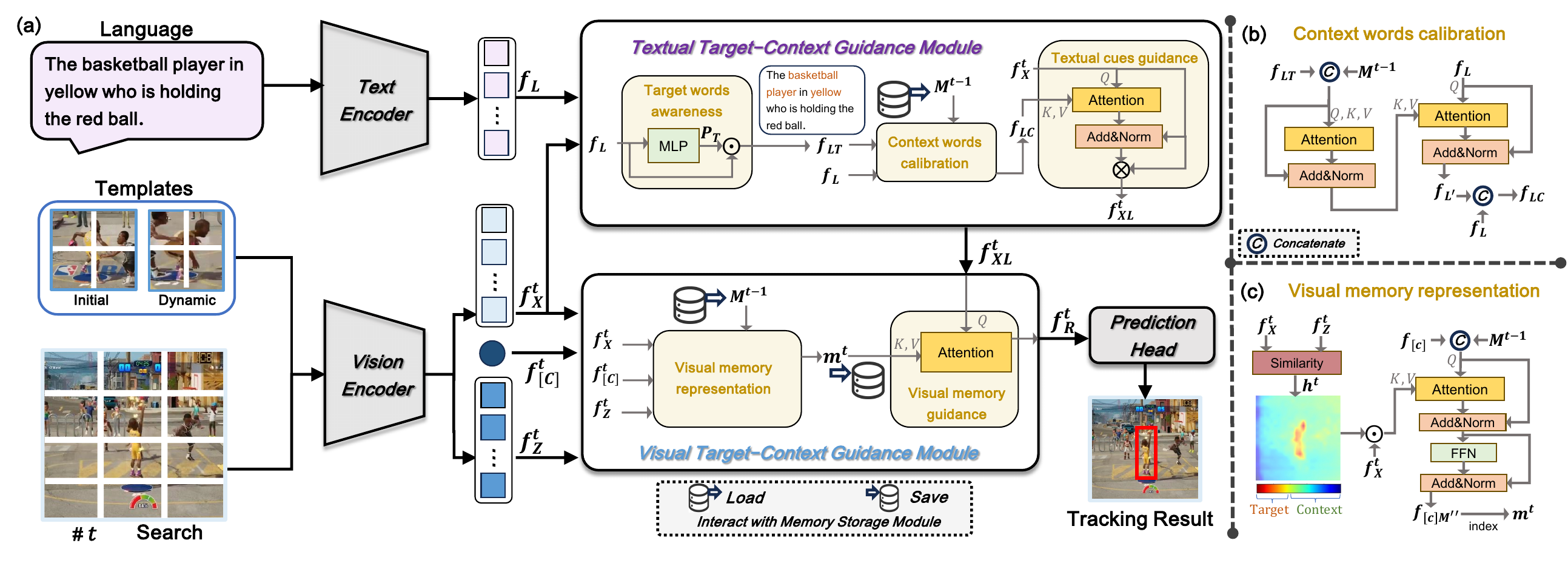}
    \caption{
    \textbf{(a) Framework of our proposed ATCTrack.} 
Given the language description and template patches as references, ATCTrack locates the target in the search image at time $t$. The input is first encoded using \textbf{\textit{Text}} and \textbf{\textit{Vision Encoders}}. Subsequently, the \textbf{\textit{Textual and Visual Target-Context Guidance Modules}} sequentially embed the aligned textual and visual cues into the search features. During this process, the \textbf{\textit{Memory Storage Module} (MSM)} provides previously stored memory information (up to $t - 1$) and saves updated memory information. Then, the \textbf{\textit{Prediction Head}} generates the final tracking results based on the embedded search features.
\textbf{(b-c) Specific model structures for context words calibration and visual memory representation.}
    }
    \label{fig3}
    \vspace{-0.4cm} 
\end{figure*}

The framework of ATCTrack is depicted in \cref{fig3}. 
Given multimodal cues and the search image, the \textbf{Text Encoder} and \textbf{Vision Encoder} first encode them into specific feature spaces. 
Then, the \textbf{Textual Target-Context Guidance Module} and \textbf{Visual Target-Context Guidance Module} sequentially model comprehensive textual and visual target-context features aligned with the dynamic search target, and embed them into the search features.
During this process, the \textbf{Memory Storage Module} (\textbf{MSM}) provides stored visual memory features and saves the updated memory information. 
Finally, the  \textbf{Prediction Head} is used to obtain the tracking result based on embedded search features.
In the following sections, we will introduce each module in detail.

\subsection{Input Encoder\label{sec:input_encoder}}
\paragraph{Vision encoder.} At time step \( t \), our visual input consists of a search image \( x_t \in \mathbb{R}^{3 \times H_x \times W_x} \), an initial template patch \( z_0 \in \mathbb{R}^{3 \times H_z \times W_z} \), and a dynamic template patch \( z_t \in \mathbb{R}^{3 \times H_z \times W_z} \) updated based on the tracking results \cite{yan2021learning,wang2024temporal}.
We adopt the one-stream network encoding paradigm \cite{OSTrack}, which has been widely used by recent mainstream trackers \cite{zhou2023joint,zheng2023towards,hong2024onetracker}. 
Specifically, \( x_t \), \( z_0 \), and \( z_t \) are first projected into token embeddings. Furthermore, inspired by MemVLT \cite{feng2025memvlt}, we introduce a learnable [CLS] token to capture global visual semantic features \cite{devlin2018bert}. This token is concatenated with template-search tokens and fed into transformer layers for feature extraction and relational modeling. 
Finally, we obtain the encoded search feature $ f_{X}^t \in \mathbb{R}^{N_x \times D} $, template feature $ f_{Z}^t \in \mathbb{R}^{N_z \times D} $ corresponding to two template images, and the aggregated [CLS] token $ f^t_{[C]} \in \mathbb{R}^{1 \times D} $.

\paragraph{Text encoder.} We utilize RoBERTa \cite{devlin2018bert,liu2019roberta}, a classic pre-trained model, as our text encoder. 
Specifically, for a given sentence, we tokenize it into a sequence of text tokens. The token sequences then are fed into the transformer layers to extract the text embedding feature \(f_L \in \mathbb{R}^{N_l \times D}\).

\subsection{Textual Target-Context Guidance Module \label{module_text}}
To fully leverage textual cues, the key insight of this module is to carefully modulate the initial text prompt to adequately focus on the target words and mitigate the potential misleading effects of context words. As shown in \cref{fig3}, we first explicitly identify the target words from the text content. 
Then, we design a context words calibration mechanism that leverages the identified target words and the visual memory features provided by the Memory Storage Module (\textbf{MSM}) to effectively utilize context words. 
Finally, we integrate the modulated text features with the search features to achieve the guidance of the textual cues.

\paragraph{Target words awareness.}
Compared to context words, the information in target words is usually constant and can serve as consistent cues for tracking. However, the diverse nature of textual expressions makes accurately identifying target words challenging.
Existing VLTs, such as QueryNLT \cite{shao2024queryvlt} and  TTCTrack \cite{mao2024textual}, attempt to distinguish them based on vision-text similarity. 
This requirement for fine-grained multimodal alignment increases the learning difficulty.
Intuitively, we can identify the target words solely based on the text content. Therefore, we simplify the identification of target words into a multi-label binary classification task, determining whether each word in the sentence belongs to the target words.

Considering the lack of relevant label in existing benchmarks \cite{hu2023multi,LaSOT,TNL2K,OTB-Lang}, we design specific prompts and leverage the powerful text comprehension capabilities of LLMs \cite{touvron2023llama,bai2023qwen} to construct an automated and reliable target words annotation pipeline.
Due to space constraints, the detailed construction process is described in Appendix A.
Leveraging this tailored target words labeling information, we find that accurate target words awareness can be achieved using a lightweight multilayer perceptron built upon the initial text features \(f_L\).
\begin{equation}
p_{T} = MLP(f_L),
\label{equ1}
\end{equation}
where \( p_{T} \in [0, 1]^{N_l \times 1} \) denotes the probability of each text token being a target word. By weighting \(f_L\) with \( p_{T} \), we obtain the target words feature \(f_{LT} \in \mathbb{R}^{N_l \times D} \). 

\paragraph{Context words calibration.} 
The alignment between context words and target states determines whether they guide or mislead the tracker. An intuitive way to determine the timing for utilizing context words is to assess them based on the dynamic evolution of the target states \cite{feng2025memvlt,shao2024queryvlt}. 
Therefore, this modeling process requires the temporal memory features stored by the MSM, which represent the latest visual target-context information (details will be described later). 
Additionally, since context information depends on the target's location, accurately perceiving the target features helps to capture precise context information \cite{bar2004visual,castelhano2010relative}. Based on these insights, the core of our context words calibration mechanism is to first enhance the target features' representation by integrating identified target words and visual temporal features. These enhanced features are then used to modulate the initial text features to adaptively utilize the context words.

At the current time $ t $, the MSM stores $ L_m $ memory units up to time step $ t-1 $, denoted as \( M^{t-1} = \left\{ m_i \right\}_{i=1}^{L_m} \).
As shown in \cref{fig3} (b), we first enhance the target feature representation by leveraging the complementarity between the visual memory features $ M^{t-1} $ and the textual target words $ f_{LT} $. Specifically, we concatenate the two features:
\begin{equation}
   f_{LM} = [ f_{LT} ; M^{t-1}_{*} ].
   \label{equ:l1}
\end{equation}
Where [;] denotes concatenation along the first dimension, and $ M^{t-1}_{*} \in \mathbb{R}^{L_m \times D} $ is the concatenation of elements in $ M^{t-1} $. Considering the transformer layers' exceptional ability in modeling feature interactions, we apply the vanilla transformer attention operation \cite{Transformer} to $ f_{LM} $:
\begin{equation}
    f_{LM^{\prime}} = Norm(f_{LM} + \Phi_{CA}(f_{LM}, f_{LM})).
    \label{equ:l2}
\end{equation}
Here, \( \Phi_{CA}(\cdot, \cdot) \) denotes the cross-attention operation, where the first element serves as the query \(Q\) and the second element is used to obtain the key \(K\) and value \(V\) \cite{Transformer}. \( Norm \) represents the layer normalization operation.

With the enhancement of target features, the surrounding context information is also indirectly perceived more accurately. Therefore, we use $ f_{LM^{\prime}} $ to modulate the initial static text $ f_{L} $, allowing for adaptive calibration of the context words.
Through the following attention operation, we obtain the meticulously calibrated text features \( f_{L^{\prime}} \in \mathbb{R}^{N_l \times D} \):
\begin{equation}
f_{L^{\prime}} = Norm( f_L + \Phi_{CA}(f_L, f_{LM^{\prime}})).
\label{equ:l4}
\end{equation}

\paragraph{Textual cues guidance.}
Inspired by recent generation models \cite{esser2024scaling,ma2025step,kong2024hunyuanvideo} that concatenate various types of text features for utilization, we treat the initial text and the calibrated text as two different types and concatenate them to obtain the comprehensive text feature $ f_{LC} = [ f_{L} ; f_{L^{\prime}} ]$.

Then, we employ transformer-based cross-attention operations and residual multiplication \cite{zheng2023towards} to obtain the visual search features $ f^t_{XL} \in \mathbb{R}^{N_x \times D} $ embedded with textual cues.

\begin{table*}[t]
    \vspace{-0.5cm} 
    \centering
    \begin{tabular}{l|ccc|ccc|ccc|ccc}
    \toprule
     \multicolumn{1}{c|}{\multirow{2}{*}{Method}}
      & \multicolumn{3}{c|}{MGIT (Action)} & \multicolumn{3}{c|}{TNL2K} &\multicolumn{3}{c|}{LaSOT} & \multicolumn{3}{c}{LaSOT\(_{ext}\)} \\ \cline{2-13}
     & AUC & P${_{\text{Norm}}}$ & P & AUC & P${_{\text{Norm}}}$ & P & AUC & P${_{\text{Norm}}}$ & P & AUC & P${_{\text{Norm}}}$  &P  \\
      \bottomrule
      \multicolumn{5}{l}{\textit{Basic Variants}} \\
     \midrule
      JointNLT \protect\cite{zhou2023joint} & 61.0 & 78.6 & 44.5 & 56.9 & 73.6 & 58.1 & 60.4 & 69.4 & 63.6 & - & - & - \\
      DecoupleTNL \protect\cite{ma2023tracking} & - & - & - & {56.7} & {-} & {56.0} & 71.2 & {-} &{75.3} & -& - & - \\
      All-in-One \protect\cite{zhang2023all} & - & - & - & 55.3 & - & 57.2 & 71.7 & 82.4 & 78.5 & 54.5 & 63.5 & - \\ 
      MMTrack \protect\cite{zheng2023towards} & - & - & - & 58.6 & 75.2 & 59.4 & {70.0} & 82.3 & 75.7& 49.4 & 59.9 & 55.3 \\ 
      QueryNLT \protect\cite{shao2024queryvlt} & - & - & - & {56.9} & {73.6} & {58.1} & {59.9} & {69.6} &{63.5} & - & - & -\\ 
      OneTracker \protect\cite{hong2024onetracker} & - & - & - & {58.0} & {-} & {59.1} & {70.5} & {79.9} & 76.5 & - & - & -\\ 
      UVLTrack-B \protect\cite{ma2024unifying} & - & - & - & 62.7 & - & 65.4 & 69.4 & - & 74.9 &  49.2 & - & 55.8 \\ 
      CTVLT \protect\cite{feng2024enhancing} & 69.2 & - & 62.9 & 62.2 & - & 79.5 & 72.3 & - & 79.7 & - & - & - \\ 
      ChatTracker-B \protect\cite{sun2025chattracker} & - & - & - & 59.6 & 76.3 & 62.1 & 71.7 & 80.9 & 77.5 & - & - & - \\ 
      MemVLT \protect\cite{feng2025memvlt} & {\color{blue}69.4} & {\color{blue}81.3} & {\color{blue}63.7} & 63.3 & {\color{blue}80.9} & 67.4 & 72.9 & 85.7 & 80.5 & 52.1 & 63.3 & 59.8 \\
      SUTrack-B224 \protect\cite{chen2024sutrack} & - & - & - & 65.0 & - & 67.9 & 73.2 & 83.4 & 80.5 & {\color{blue}53.1} & {\color{blue}64.2} & {\color{blue}60.5} \\ 
      SUTrack-B384 \protect\cite{chen2024sutrack} & - & - & - & {\color{blue}65.6} & - & {\color{blue}69.3} & {\color{blue}74.4} & {\color{blue}83.9} & {\color{blue}81.9} & 52.9 & 63.6 & 60.1 \\ 
      \midrule[0.1pt]
      \rowcolor{gray!20}
      \textbf{ATCTrack-B} & {\color{red}73.7} &  {\color{red}84.5} & {\color{red}70.1} & {\color{red}67.5} & {\color{red}85.3} & {\color{red}73.6} & {\color{red}74.6} & {\color{red}87.0} &{\color{red}82.1} & {\color{red}54.6} & {\color{red}65.7} & {\color{red}62.8} \\ 
    \bottomrule
     \multicolumn{5}{l}{\textit{Performance-oriented Variants}} \\
     \midrule
     ChatTracker-L \protect\cite{sun2025chattracker} & - & - & - & 65.4 & {\color{blue}76.5} & 70.2 & 74.1 & 83.8 & 81.2 & - & - & - \\
     UVLTrack-L \protect\cite{ma2024unifying} & - & - & - & 64.8 & - & 68.8 & 71.3 & - & 78.3 &  51.2 & - & 59.0 \\ 
     SUTrack-L224 \protect\cite{chen2024sutrack} & - & - & - & 66.7 & - & 70.3 & 73.5 & 83.3 & 80.9 & {\color{blue}54.0} & {\color{blue}65.3} & {\color{blue}61.7} \\ 
     SUTrack-L384 \protect\cite{chen2024sutrack} & - & - & - & {\color{blue}67.9} & - & {\color{blue}72.1} & {\color{red}75.2} & {\color{blue}84.9} & {\color{red}83.2} & 53.6 & 64.2 & 60.5 \\ 
     \midrule[0.1pt]
      \rowcolor{gray!20}
      \textbf{ATCTrack-L} & {\color{red}74.0} &  {\color{red}86.5} & {\color{red}76.1} & {\color{red}68.6} & {\color{red}85.8} & {\color{red}75.0} & {\color{blue}74.7} & {\color{red}87.1} &{\color{blue}82.3} & {\color{red}55.4} & {\color{red}66.8} & {\color{red}64.0} \\
     \bottomrule
    \end{tabular} 
    \caption{Comparison with state-of-the-art trackers on four popular benchmarks: MGIT \protect\cite{hu2023multi}, TNL2K \protect\cite{TNL2K}, LaSOT \protect\cite{LaSOT}, and LaSOT\(_{ext}\) \cite{fan2021lasot}. 
    The best two results are highlighted in {\color{red}red} and {\color{blue}blue}, respectively.}
    \label{tab:results_sota}
    \vspace{-0.3cm} 
\end{table*}

\subsection{Visual Target-Context Guidance Module  \label{module_vision}}
As shown in \cref{fig3}, this module consists of two core processes: visual memory representation and guidance, aiming to model and leverage dynamic visual target-context memories aligned with the target states.  
Compared with the sparse dynamic template \cite{STARK}, visual memories can provide denser temporal features. These two mechanisms jointly offer comprehensive temporal information for the tracker.
For visual memory representation, we explicitly construct the target-context distribution map and model the memory features at different time steps.  
Subsequently, we embed the memory cues stored across multiple time steps into the search features, thereby guiding the tracking process.

\paragraph{Visual memory representation.} 
For the [CLS] token \( f^t_{[C]} \) encoded by the vision encoder, since it participates in the entire feature integration process of visual information, it serves as a suitable global visual feature representation \cite{devlin2018bert,feng2025memvlt,zhang2025beyond}. 
Thus, an intuitive approach is to treat \( f^t_{[C]} \) as the memory representation at the current time step.
However, since the vision encoder is part of the model's early feature modeling stage \cite{OSTrack}, the precise target location may not have been sufficiently perceived, meaning that \( f^t_{[C]} \) lacks explicit awareness of the target-context distribution information. 
Therefore, we attempt to explicitly construct the target-context distribution heatmap and embed it into \( f^t_{[C]} \) to obtain our visual memory.

Specifically, we construct the target-context distribution heatmap by calculating the similarity between the encoded search feature \( f_X^t \) and the template features \( f_{Z}^t \) \cite{OSTrack}:
\begin{equation}  
     h^t = (f_X^t \cdot (f_Z^t )^T).mean(dim=1).  
\end{equation}  
Since the \( f_{Z}^t \) are centered on the target, \( h^t \in \mathbb{R}^{N_x \times 1} \) reflects the probability of each search token belonging to the target (\ie, not being part of the context).  
The heatmaps in \cref{fig3} (c) are obtained by reshaping \( h^t \) into a two-dimensional image space, demonstrating its high interpretability in indicating the spatial distribution of the target-context.
 
Next, $h^t$ is embedded into $f^t_{[C]}$ to construct the current memory unit $m^t$. 
To provide richer target-context dynamic information when constructing $m^t$, we introduce the historical memory $M^{t-1}$ stored in MSM up to step $t-1$. 
Note that the $L_m$ memory units in $M$ are obtained through iterative storage of $m^t$ at different time steps. 
Assuming $M^{t-1}$ has been obtained, we will elaborate on the generation process of each memory unit in the following, while the updating mechanism of $M$ will be presented in \cref{MSM}.

For implementation, we employ the vanilla cross-attention mechanism \cite{Transformer}.
In particular, we concatenate the features of $f^t_{[C]}$ and $M_{*}^{t-1}$ to obtain $f_{[C]M}$ as the query, while using the $h^t$-weighted $f_x^t$ as both key and value:
\begin{equation}  
     f_{[C]M^{\prime}}=  Norm(f_{[C]M}+ \Phi_{CA}(f_{[C]M}, h^t\odot f^t_X )),  
    \label{equ_vem}  
\end{equation}  
\begin{equation}  
    f_{[C]M^{\prime\prime}}=  Norm( f_{[C]M^{\prime}}+ FFN(f_{[C]M^{\prime}})).  
    \label{equ:l3}  
\end{equation}  
Where $\odot$ represents the Hadamard product, and $FFN$ denotes the feed-forward network. 
Based on the concatenation index, we extract the feature corresponding to $f^t_{[C]}$ from $f_{[C]M^{\prime\prime}}$ and utilize it as the memory unit $m^t$ for the current time step. 
This establishes our memory unit generation mechanism. The generated $m^t$ is then stored in the MSM for subsequent time step computations (see \cref{MSM}).

\paragraph{Visual memory guidance.} 
For the $m^t$ that incorporates both current and historical target-context feature information, we adopt parameter-free attention operations \cite{xie2024autoregressive} to facilitate its interaction with the search feature $f^t_{XL}$:
\begin{align}
    f^t_{R} &=  softmax(\frac{f^t_{XL} \cdot (m^t)^T}{\sqrt{D}})\cdot m^t,
\end{align}
Here, $f^t_{R} \in \mathbb{R}^{N_x \times D}$ denotes the search feature that incorporates both textual and visual memory cues, which is subsequently fed into the prediction head.

\subsection{Memory Storage Module \label{MSM}}
As previously mentioned, this module provides the tracker (at \( t \)) with stored memory features $M^{t-1}$ (up to \( t-1 \)). Meanwhile, it stores the newly generated memory unit $m^t$ for guiding the tracking process at the next time step ($t+1$). 
The $L_m$ memory units in $M$ are initialized with $f^0_{[C]}$ from the first frame and updated using a simple yet widely adopted sliding window approach \cite{cai2024hiptrack,xie2024autoregressive}. 
For further details, please refer to Appendix C.2.

\subsection{Prediction Head and Loss}
Based on the search feature \( f^t_{R} \)  which integrates multimodal cues, we utilize a classic CNN-based prediction head \cite{OSTrack,xie2024autoregressive} to obtain the final bounding box.
We employ the focal loss \(L_{cls}\) \cite{law2018cornernet}, \(L_1\) loss, and the generalized IoU loss \(L_{iou}\) \cite{rezatofighi2019generalized} to supervise the prediction of bounding box, which are widely used in tracker design\cite{OSTrack}. 
Additionally, for the target words classification task, we employ the binary cross-entropy loss \(L_{bce}\) for supervision.
The overall loss function is formulated as follows:
\begin{equation}
    L_{all}= L_{cls} + 2\times L_{iou} + 5\times L_1 + L_2 + 0.2\times L_{bce}.
\end{equation}

\section{Experiments}
\subsection{Implementation Details\label{imp_detail}}
We adopt RoBERTa-Base \cite{liu2019roberta} as our text encoder and, following recent advanced trackers \cite{shi2024explicit,xie2024autoregressive,chen2024sutrack}, employ HiViT \cite{zhang2022hivit,tian2024fast} as our vision encoder. 
To balance performance and computational efficiency, we develop two model variants: ATCTrack-B and ATCTrack-L, initialized with Fast-iTPN-B and Fast-iTPN-L \cite{tian2024fast} respectively, with token dimensions $D$ of 512 and 768. 
The template patches and search images are sized at $128 \times 128$ and $256 \times 256$, respectively.
Additionally, the default length of MSM is set to four, and the dynamic template update strategy follows the STARK \cite{STARK}. 
Our tracker is trained on a server with four A5000 GPUs and tested on an RTX-3090 GPU.
The tracking speed of ATCTrack-B/L is 35/30 FPS. 
For a comparison of our model with mainstream VLTs in terms of parameters and speed, please refer to  Appendix F.1.

We train our model using the training splits from LaSOT \cite{LaSOT}, TNL2K \cite{TNL2K}, RefCOCOg \cite{mao2016generation}, OTB99-Lang \cite{OTB-Lang}, VastTrack \cite{peng2025vasttrack}, GOT-10k \cite{huang2019got}, and TrackingNet \cite{muller2018trackingnet}. For GOT-10k \cite{huang2019got} and TrackingNet \cite{muller2018trackingnet} that lack text annotations, we follow the All-in-One's strategy \cite{zhang2023all} by treating class names as pseudo language labels. 
Each training sample comprises a text description, two template patches, and four search frames from the same video sequence. Our tracker performs iterative training by sequentially selecting search images.
The network parameters are optimized using AdamW optimizer \cite{loshchilov2017decoupled} for 150 epochs, with each epoch containing 20,000 randomly sampled instances. 

\subsection{Comparison with State-of-the-arts\label{exp:sota_compare}}

\noindent\textbf{MGIT.} 
MGIT is a latest VLT benchmark specifically tailored for the complex long-term scenarios.
Each sequence contains challenging spatio-temporal causal relationships and is annotated with text prompts at three levels of granularity \cite{hu2022global,hu2023multi,li2024visual,li2024dtvlt,li2024texts,feng2025narrlv}. 
As shown in \cref{tab:results_sota}, ATCTrack demonstrates superior performance at the representative action granularity.
Particularly, ATCTrack-B surpassing the SOTA tracker MemVLT \cite{feng2025memvlt} by 4.3\%, 3.2\%, and 6.4\% in area under the curve (AUC), normalized precision (P\(_\text{Norm}\)), and precision score (P), respectively.
Unlike MemVLT's implicit modulation of static multimodal cues, ATCTrack explicitly adapts these cues from a target-context perspective. These results validate our approach's effectiveness in handling complex long-term scenarios.

\noindent\textbf{TNL2K.} 
TNL2K is also designed for the VLT task, and the introduction of attributes such as “adversarial samples" and “modality switch" significantly adds to the challenges \cite{TNL2K}.
As shown in \cref{tab:results_sota}, ATCTrack-L outperforms the recent ChatTracker-L by 4.8\% in P. 
Compared to  ChatTracker employs multimodal large language models to generate high-quality text annotations for addressing static text limitations, our textual target-context guidance module achieves superior performance with fewer network parameters.

\noindent\textbf{LaSOT \& LaSOT\(_{ext}\).} 
They are extensions of traditional visual tracking benchmarks \cite{LaSOT,OTB2015} by adding text labels, focusing on long-term tracking challenges. 
Furthermore, LaSOT\(_{ext}\) also includes many similar distractors, further complicating the tracking task.
As shown in \cref{tab:results_sota}, except for marginally lower AUC and P scores compared to SUTrack-L384, which utilizes larger image resolution and more training datasets, ATCTrack demonstrates highly competitive performance across all other metrics. 
The outstanding performance across multiple benchmarks further reflects ATCTrack's strong generalization capability.

\begin{figure}[t!]
    \centering
    \includegraphics[width=\linewidth]{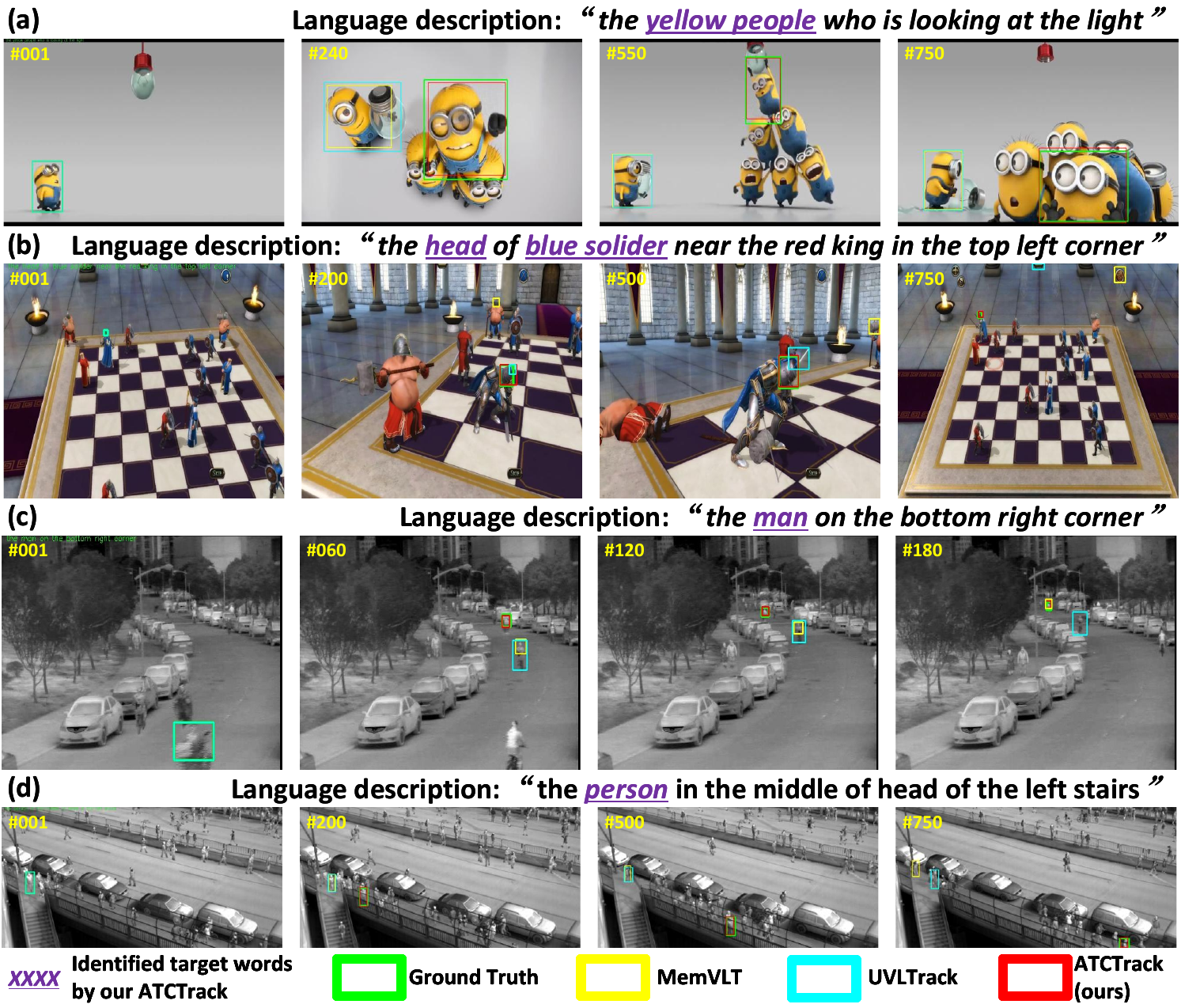}
    \caption{
    Qualitative comparison results of our tracker with other two VLTs (\ie, MemVLT and UVLTrack ) on four challenging cases. Better viewed in color with zoom-in.
    }
    \label{fig4}
\end{figure}

\noindent\textbf{Qualitative comparison.} 
As shown in \cref{fig4}, we present the tracking results of ATCTrack-B and two existing SOTA VLTs\cite{feng2025memvlt,ma2024unifying} on four challenging sequences.
In these cases, the appearance of the target undergoes significant changes \cite{feng2025cstrack,ling2025vmbench}, and the language descriptions contain context words that may lead to ambiguity.
It is evident that our ATCTrack exhibits greater robustness \cite{chen2024revealing} and effectiveness.
More cases can be found in Appendix G.

 \subsection{Ablation Study \label{sec_ab}}
To investigate the properties of the various modules in ATCTrack, we conduct comprehensive ablation studies on ATCTrack-B using TNL2K \cite{TNL2K} and LaSOT \cite{LaSOT}.
For the implementation details of each setting, see Appendix E.

\begin{table}[t]
    \centering
    \resizebox{1\linewidth}{!}{
        \setlength{\tabcolsep}{2mm}{
            \small
            \begin{tabular}{c|cc|cc|cc}
                \toprule
                \multirow{2}*{\#} & \multicolumn{2}{c|}{Setting} & \multicolumn{2}{c|}{TNL2K} & \multicolumn{2}{c}{LaSOT} \\
                \cline{2-7}
                 & \textit{Textual\(_{TC}\)} & \textit{Visual\(_{TC}\)} & AUC & P & AUC & P \\
                \midrule[0.5pt]
                1 &  &  & 65.5 & 70.6 & 72.5 & 79.8 \\
                2 & \ding{51} &  & 66.5 & 72.0 & 73.6 & 81.3 \\
                3 &  & \ding{51} & 66.7 & 72.3 & 73.7 & 81.7 \\
                4 & \ding{51} & \ding{51} & {\color{red}67.5} & {\color{red}73.6} & {\color{red}74.7} & {\color{red}82.3} \\
                \bottomrule
            \end{tabular}
        }
    }
    \caption{Ablation study on important model components.}
    \label{tab:ab_1}
\end{table}

\noindent\textbf{Study on important model components.} 
The core contribution of our work is the introduction of a novel multimodal target-context modeling mechanism.
In \cref{tab:ab_1} (\#1), we show results using only the initial prompts and the dynamic template to guide tracker. \cref{tab:ab_1} (\#2) and (\#3) present results using our textual and visual target-context guidance modules, respectively.
Comparatively, Our method shows superior performance with 1.0\% and 1.2\% AUC gains on TNL2K, respectively, demonstrating the effectiveness of our approach. 
Additionally, \cref{tab:ab_1} (\#4) indicates that combining these two modules provides complementary benefits, further enhancing tracking performance.

\noindent\textbf{Study on textual target-context modeling.}
Similar to most VLTs \cite{GTI,zhou2023joint,zheng2023towards}, \cref{tab:ab_2} (\#1) (identical to \cref{tab:ab_1} (\#1)) treats the text features as a whole, indiscriminately fusing them with visual features. 
\cref{tab:ab_2} (\#2) and (\#3) successively incorporate the target words awareness and context words calibration mechanisms to model and leverage the textual target-context cues aligned with the target state.
The progressive improvement in performance demonstrates the effectiveness of our proposed textual guidance approach.

Besides, we utilize both the initial and calibrated textual feature $f_{LC}$ for tracking guidance. 
\cref{tab:ab_2} (\#4) presents the results of using only the calibrated textual feature $f_{L^{\prime}}$. 
Compared to \cref{tab:ab_2} (\#3), the degraded model performance indicates that dual-type textual features help the tracker leverage more comprehensive information.

Moreover, we quantitatively evaluate the target words identification accuracy of our method compared to the existing vision-text similarity-based method (see Appendix B for details). 
As shown in \cref{fig2} (a), despite using only a lightweight multilayer perceptron (\cref{equ1}), our method achieves impressive accuracy, distinguishing target words with an overall accuracy (Acc\(_\text{all}\)) of 98.9 \%.
This facilitates the tracker's utilization of textual target-context cues.
Furthermore, our experiments reveal that lightweight text analysis tools, such as Scene Graph Parse \cite{schuster2015generating}, demonstrate poor performance in target word recognition, achieving an accuracy ($\text{Acc}_{\text{target}}$) of only 21.0\%.

\begin{table}[t]
    \centering
    \resizebox{1\linewidth}{!}{
        \setlength{\tabcolsep}{2mm}{
            \small
            \begin{tabular}{l|l|cc|cc}
                \toprule
                \multirow{2}*{\#} & \multirow{2}*{Setting} & \multicolumn{2}{c|}{TNL2K} & \multicolumn{2}{c}{LaSOT} \\
                \cline{3-6}
                 &  & AUC & P & AUC & P \\
                \midrule[0.5pt]
                1 & \textit{naive method} & 65.5 & 70.6 & 72.5 & 79.8 \\
                2 & \textit{+ target words awareness} & 65.9 & 71.7 & 72.9 & 80.4 \\
                3 & \textit{+ context words calibration} & {\color{red}66.5} & {\color{red}72.0} & {\color{red}73.6} & {\color{red}81.3} \\
                4 & \textit{- dual-type textual guidance} & 66.2 & 71.8 & 73.2 & 80.3 \\
                \bottomrule
            \end{tabular}
        }
    }
    \caption{Ablation study on textual target-context modeling.}
    \label{tab:ab_2}
\end{table}

\noindent\textbf{Study on visual target-context modeling.}
\cref{tab:ab_3} (\#1) shows the results when only sparse initial and dynamic template patches are employed. 
To model and leverage denser dynamic target-context features, we propose our visual target-context guidance module. 
\cref{tab:ab_3} (\#3, \#4, \#5) explores different implementation approaches. 
\cref{tab:ab_3} (\#3) adopts RoI processing~\cite{wang2021transformer}, a classic local image feature modeling method, which replaces $h^t\odot f^t_X$ in \cref{equ_vem} with the local region in $f^t_X$ determined by the predicted bbox ($\times1.5$). 
Similarly, \cref{tab:ab_3} (\#4) replaces $h^t$ with the local mask of predicted bbox ($\times1.5$). 
Compared to \cref{tab:ab_3} (\#2), these two methods achieve slight improvements on TNL2K but suffer significant performance drops on LaSOT. 
In contrast, our method (\ie, \cref{tab:ab_3} (\#5)) achieves superior results, demonstrating the necessity of representing target-context information from a global perspective.

\begin{table}[t]
    \centering
    \resizebox{1\linewidth}{!}{
        \setlength{\tabcolsep}{2mm}{
            \small
            \begin{tabular}{l|l|cc|cc}
                \toprule
                \multirow{2}*{\#} & \multirow{2}*{Setting} & \multicolumn{2}{c|}{TNL2K} & \multicolumn{2}{c}{LaSOT} \\
                \cline{3-6}
                 &  & AUC & P & AUC & P \\
                \midrule[0.5pt]
                1 & \textit{naive method} & 65.5 & 70.6 & 72.5 & 79.8 \\
                2 & \emph{w} \textit{RoI} & 65.6 & 71.1 & 71.1 & 77.4 \\
                3 & \emph{w} \textit{search + crop mask} & 65.9 & 71.4 & 71.9 & 78.7 \\
                4 & \emph{w} \textit{search + global mask} & {\color{red}66.7} & {\color{red}72.3} & {\color{red}73.7} & {\color{red}81.7} \\
                \bottomrule
            \end{tabular}
        }
    }
    \caption{Ablation study on visual target-context modeling.}
    \label{tab:ab_3}
\end{table}

\begin{table}[t]
    \centering
    \resizebox{1\linewidth}{!}{
        \setlength{\tabcolsep}{2mm}{
            \small
            \begin{tabular}{l|l|cc|cc}
                \toprule
                \multirow{2}*{\#} & \multirow{2}*{Setting} & \multicolumn{2}{c|}{TNL2K} & \multicolumn{2}{c}{LaSOT} \\
                \cline{3-6}
                 &  & AUC & P & AUC & P \\
                \midrule[0.5pt]
                1 & \textit{ATCTrack-B} & 67.5 & 73.6 & 74.7 & 82.3 \\
                2 & \emph{w/o} \textit{HiViT backbone } & 65.8 ($\downarrow$ 1.7) & 71.3 ($\downarrow$ 2.3) & 72.9 ($\downarrow$ 1.8) & 80.7 ($\downarrow$ 1.6) \\
                3 & \emph{w/o} \textit{dynamic template} & 67.2 ($\downarrow$ 0.3) & 73.0 ($\downarrow$ 0.6) & 73.2 ($\downarrow$ 1.5) & 80.8 ($\downarrow$ 1.5)\\
                4 & \emph{w/o} \textit{\textit{Textual\(_{TC}\)}\&\textit{Visual\(_{TC}\)}} & 65.5 ($\downarrow$ {\color{red}2.0}) & 70.6 ($\downarrow$ {\color{red}3.0}) & 72.5 ($\downarrow$ {\color{red}2.2}) & 79.8 ($\downarrow$ {\color{red}2.5})\\
                5 & \emph{w/o} \textit{target words label} & 67.0 ($\downarrow$ 0.5) & 72.9 ($\downarrow$ 0.7) & 73.5 ($\downarrow$ 1.2) & 80.6 ($\downarrow$ 1.7)\\
                \bottomrule
            \end{tabular}
        }
    }
    \caption{Ablation study on the contribution of different modules.}
    \label{tab:ab_5}
    \vspace{-0.4cm} 
\end{table}

\noindent\textbf{Further ablation studies.} 
Several components in our model, such as the HiViT backbone \cite{zhang2022hivit,tian2024fast} and dynamic template \cite{STARK}, are well-established designs for improving tracking performance. 
Therefore, it is necessary to assess their impact relative to our core contribution, \ie, the multimodal target-context guidance modules. 
As shown in \cref{tab:ab_5} (\#2, \#3), replacing HiViT with ViT \cite{OSTrack} or removing the dynamic template leads to performance degradation, which aligns with existing findings\cite{xie2024autoregressive,STARK}. 
However, \cref{tab:ab_5} (\#4) reveals that removing our multimodal target-context guidance module results in more significant performance drops, indicating that ATCTrack's superior performance primarily stems from our proposed method.

Furthermore, ATCTrack employs unique target word supervision labels, which constitute a significant contribution of our method. To ensure fairness, \cref{tab:ab_5} (\#5) presents the results without this label. Although the performance of the model decreases, it still surpasses the SOTA trackers represented by MemVLT \cite{feng2025memvlt} and SUTrack-B224 \cite{chen2024sutrack}.

\section{Conclusion}
In pursuit of robust vision-language tracking, especially in complex  long-term scenarios that reflect real-world conditions as exemplified by MGIT, we propose a novel tracker named ATCTrack.
Through comprehensive target-context modeling, we obtain multimodal cues aligned with dynamic target states, offering an innovative solution to the obstacle where initial static cues fail to provide sustained guidance.
For the textual modality, we introduce a precise target words awareness method to ensure sufficient attention to target words, and design an innovative context words calibration mechanism to mitigate potential misleading effects. 
For the visual modality, we efficiently characterize temporal target-context features to provide timely visual cues for tracking.
Through these combined efforts, our model achieves outstanding performance, significantly surpassing existing methods across four mainstream benchmarks.

\section*{Acknowledgements}
This work is supported in part by the National Natural Science Foundation of China (Grant No.62176255).

{
    \small
    \bibliographystyle{ieeenat_fullname}
    \bibliography{main}
}
\clearpage
\setcounter{page}{1}
\maketitlesupplementary
\renewcommand{\thesection}{\Alph{section}}
\renewcommand{\theequation}{A\arabic{equation}} 
\setcounter{equation}{0}
\setcounter{section}{0}

\section{Target Words Annotation Pipeline \label{pipeline}}
Given the inherently flexible and diverse nature of textual descriptions, it is challenging for trackers to accurately identify target words and context words. In our work, we approach the identification of target words as a multi-label binary classification task, enhancing the model's ability to recognize target words through supervised learning. However, existing benchmarks \cite{hu2023multi,TNL2K,fan2021lasot,peng2025vasttrack} provide only textual descriptions without labeled information on the types of target words (\ie, target words or context words). 
For such a natural language processing task, we leverage the powerful text understanding capabilities of the large language models \cite{touvron2023llama,hurst2024gpt} to construct an automated target words annotation pipeline. 
Specifically, we employ the widely-used multimodal large language model, GPT-4o \cite{hurst2024gpt}, and have devised a specific core prompt to guide GPT-4o in recognizing target words (as shown in \cref{fig_app1}).

Leveraging our automated annotation pipeline, we complete the labeling of target words in textual data from the MGIT \cite{hu2023multi}, TNL2K \cite{TNL2K}, LaSOT \cite{LaSOT}, RefCOCOg \cite{mao2016generation}, OTB99-Lang \cite{OTB-Lang} and Vasttrack \cite{peng2025vasttrack} datasets. We conduct a random sampling of the labeled results, inspect 50 sentences, and find that the annotations are entirely accurate. This ensures the reliability of our supervised models in classifying target words. In the future, we will open source both the target words label information and our code.

\renewcommand{\thefigure}{A\arabic{figure}}  
\setcounter{figure}{0}  

\renewcommand{\thetable}{A\arabic{table}} 
\setcounter{table}{0} 

\begin{figure*}[htbp]
    \centering
    \includegraphics[width=0.95\textwidth]{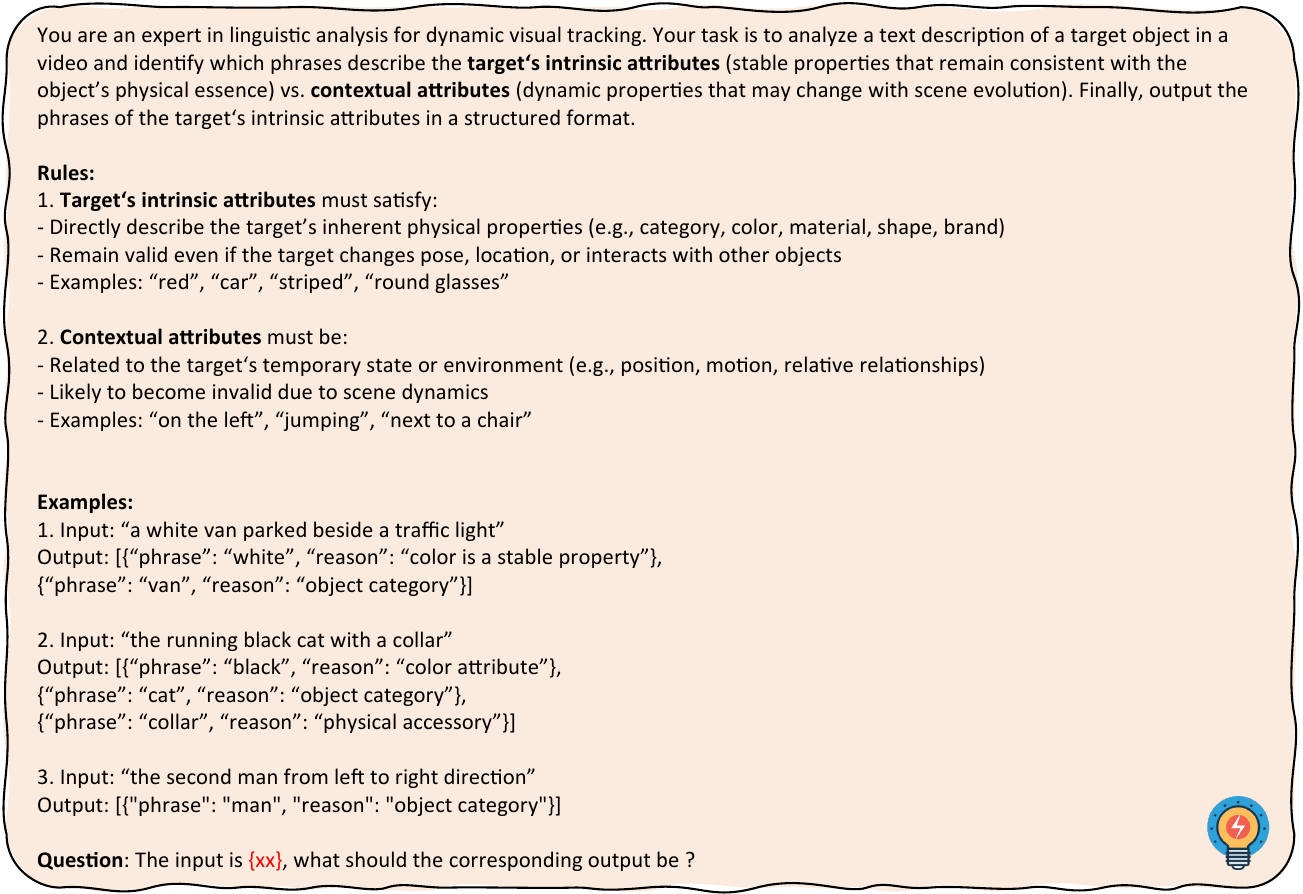}
    \caption{\textbf{Prompt used to guide GPT-4o in identifying target words information.} 
    This prompt primarily consists of two parts: task requirement descriptions and example guidance. Replace {\color{red}\{xx\}} with the sentence to be identified to achieve output results similar to the example format.
    }
    \label{fig_app1}
\end{figure*}

\begin{figure*}[htbp]
    \centering
    \includegraphics[width=0.95\textwidth]{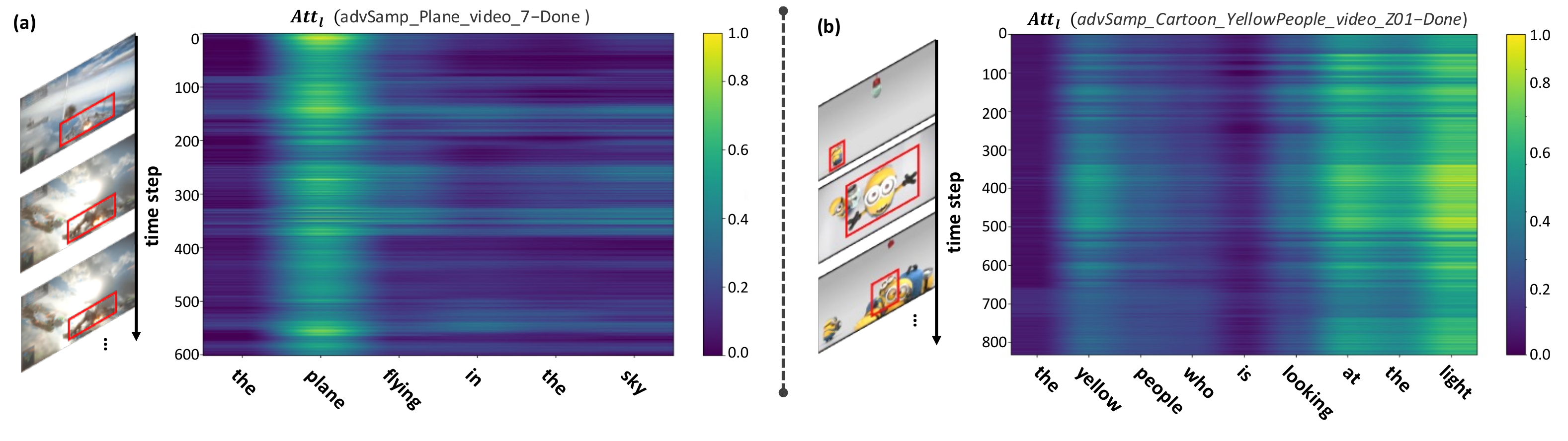}
    \caption{\textbf{Visualization results of \( Att_{l} \) across two video sequences.} 
    \textbf{(a)} In the sequence \(\text{'advSamp\_Plane\_video\_7-Done'}\), the target “plane" receives significant attention during the tracking process, which aligns with our intended effect. \textbf{(b)} In the sequence \(\text{'advSamp\_Cartoon\_YellowPeople\_video\_Z01-Done'}\), the target “yellow people" is intended to be tracked, but the tracker primarily focuses on the text “the light". This indicates that the tracker did not correctly focus on the target words, which could mislead the tracking process.
    }
    \label{fig_app2}
\end{figure*}

\section{Evaluation of Target Words Identification\label{eval_acc}}
In this section, we discuss the specific implementation methods for the target words classification accuracy results shown in Fig. 2 (a).
Recent studies, such as  QueryNLT \cite{shao2024queryvlt}, TTCTrack \cite{mao2024textual} and OSDT \cite{zhang2024one}, have utilized vision-text similarity metrics to identify target words.
Although this is one of their main contributions, they have not provided quantitative evaluation results. For this, we conduct a quantitative analysis based on the target words label information obtained from \cref{pipeline}.

\subsection{Similarity-Based Target Words Identification \label{sec_app_e1}}
Considering that QueryNLT \cite{shao2024queryvlt}, TTCTrack \cite{mao2024textual} and OSDT \cite{zhang2024one} have not open-sourced their code, we employ JointNLT \cite{zhou2023joint}, a representative vision-language tracker, as a proxy model for evaluation. The core insight of JointNLT is the use of a one-stream network to jointly model the feature extraction and interaction of text, template images, and search images. The extensive feature interaction among these elements can, to some extent, represent the feature interaction operations conducted in the aforementioned works for measuring vision-text similarity.

Specifically, at time step \(t 
 (t\geq 0)\), after the feature encoding by the JointNLT's backbone network, we obtain the visual features \( f^t_V \in \mathbb{R}^{400 \times 512} \) and the textual features \( f^t_L \in \mathbb{R}^{L \times 512} \). Here, the length of the visual tokens is fixed at 400, while the length of the textual tokens, L, is determined by the number of words in the sentence. The similarity between them is obtained through the following operations:
\begin{equation}
    att_{vl}^t = (f^t_L)^T \cdot f^t_V,
\end{equation}
where \( att_{vl}^t \in \mathbb{R}^{L \times 400} \) represents the similarity between each visual and textual token. By averaging along the dimension of the search tokens, we can determine the attention each textual token receives at the current time step \(t\), denoted as \( att_{l}^t \in \mathbb{R}^{L} \).

By concatenating \( att_{l}^t \) at each time step in a video sequence along the time dimension, we can obtain a heatmap of textual feature information for this sequence, denoted as \( Att_{l} \in \mathbb{R}^{L \times T} \), where \( T \) represents the number of frames in the video sequence.

For a more intuitive understanding, we conduct a visualization analysis using two video sequences as examples.
The related results are depicted in \cref{fig_app2}, which serves as a supplement to Fig. 2 (b) and (c) in the main text.
As shown in \cref{fig_app2} (a), the target being tracked in this sequence is “plane". In the corresponding \( Att_{l} \) heatmap, the target word “plane" receives significant attention, indicating that the tracker correctly understands the intent embedded in the text prompt, and this text cue aids in the tracking process.
For the example in \cref{fig_app2} (b), the intended tracking target is “yellow people", but the tracker primarily focuses on the word “the light". This indicates that the tracker did not correctly focus on the target words, which could mislead the tracking process.

\subsection{Evaluation of Target Words Identification Accuracy \label{sec_app_e2}}
In addition to qualitatively demonstrating the tracker's ability to distinguish each word in the text as described above, we also need to conduct a quantitative evaluation. 
First, to analyze the tracker's attention to each word throughout the entire video sequence, we average \( Att_{l} \) along the time dimension, resulting in \( Res_{l} \in \mathbb{R}^{L} \). 
Each element in \( Res_{l} \) reflects the amount of attention the tracker gives to the word at the corresponding position.

Based on this information, we can map to obtain the tracker's final prediction results for target words classification, \( p \in \{0, 1\}^L \). Specifically, based on the target words label information obtained from \cref{pipeline}, we can determine the number of target words \( k \) in the sentence. Then, we calculate the top \( k \) elements and their indices in \( Res_{l} \). Subsequently, we set the elements at these indices in \( p \) to 1, while all other elements are set to 0.

Additionally, utilizing the target words label information provided in \cref{pipeline}, we can obtain a ground truth label \( g \in \{0, 1\}^L \). In this label, 0 indicates that the word token at that position is a context word, and 1 indicates it is a target word. We then establish two accuracy assessment metrics, 
namely, Acc\(_\text{all}\) and Acc\(_\text{target}\), 
by performing different calculations on \( p \) and \( g \) to evaluate the tracker's accuracy in classifying target words.
Here, \( \text{Acc}_{\text{all}} \) represents the overall classification accuracy of the model for both target and context words;
while \( \text{Acc}_{\text{target}} \) focuses on the classification accuracy specifically for target words.

\begin{equation}
     \text{Acc}_{\text{all}} = \frac{\sum_{i=1}^L \mathbf{1}(p_i = g_i)}{L},
\end{equation}

\begin{equation}
    \text{Acc}_{\text{target}} = \frac{\sum_{i=1}^N \mathbf{1}(p_i = 1 \land g_i = 1)}{\sum_{i=1}^N \mathbf{1}(g_i = 1)}.
\end{equation}
Here, \( \mathbf{1}(\cdot) \) is an indicator function that returns 1 if the condition within the parentheses is satisfied.

Similarly, for our proposed ATCTrack and its predictions about target words \( p_{lt} \) (see Eq. (1)), we can use the same method to map it to \( p \), and then use the above formula for accuracy measurement. The corresponding accuracy results are displayed in Fig. 2 (a). It is evident that our method significantly outperforms methods based on vision-text similarity in both metrics.

\subsection{Analysis of Evaluation Results \label{sec_app_e3}}
Fig. 2 (a) shows the target words identification accuracy of our method compared to the existing vision-text similarity-based method \cite{shao2024queryvlt,mao2024textual,zhang2024one}. As can be seen, our method achieves an impressive 96.7\% in the Acc$_\text{target}$ metric, significantly surpassing the latter's 29.9\%. This precise target word awareness lays a solid foundation for subsequent text cue adjustment and utilization.
This demonstrates that our lightweight multilayer perceptron (Eq. (1)) effectively transfers the LLMs' target word distinguishing capability into the tracker. Although existing LLMs have good target word sensing capabilities, integrating LLMs directly into the tracker incurs substantial computational costs, which is detrimental to practical applications. 
Additionally, there are lightweight text component analysis tools  in the field of natural language processing, such as the widely used Scene Graph Parser \cite{schuster2015generating}. 
We evaluated the Scene Graph Parser's accuracy in identifying target words in sentences and found it to be only 21.0\%. This indicates that these tools are not yet capable of meeting our target word identification needs in a plug-and-play manner.

\section{More Details on the ATCTrack \label{model_detail}}
Due to space constraints, we focus primarily on the main contributions of our paper in the Sec. 3, specifically the textual target-context guidance module (see Sec. 3.2) and the visual target-context guidance module (see Sec. 3.3). For other components of our tracker, such as the prediction head and memory storage module, we provide a brief introduction using current mainstream methods, supplemented by relevant references. In this section, we offer an additional explanation of these components.

\subsection{Prediction Head}
The prediction head is used to predict the final bbox \(b^t\). We employ a CNN-based tracking head \cite{OSTrack,xie2024autoregressive}, which is widely adopted in tracker design. 
Firstly, for the search feature \( f^t_{R} \in \mathbb{R}^{N_x \times D} \) that integrates both textual and visual cues, we transform it into a 2D spatial feature map.
Subsequently, after passing through \(L_h\) stacked Conv-BN-ReLU layers, we obtain a classification score map \(P \in [0, 1]^{1 \times H_s \times W_s}\), the size of the bbox \(B \in [0, 1]^{2 \times H_s \times W_s}\), and the offset size \(O \in [0, 1)^{2 \times H_s \times W_s}\).
Then, the position with the highest classification score is considered to be the target position, \ie, \((x_d, y_d) = \arg\max_{(x, y)} P_{xy}\). The final target bbox is obtained as:
\begin{equation}
    x = x_d + O(0, x_d, y_d),
\end{equation}
\begin{equation}
    y = y_d + O(1, x_d, y_d),
\end{equation}
\begin{equation}
    w = S(0, x_d, y_d), 
\end{equation}
\begin{equation}
    h = S(1, x_d, y_d).
\end{equation}

\subsection{Memory Storage Module \label{msm_app}}
As introduced in Sec. 3.4, we employ the sliding windows method \cite{cai2024hiptrack, xie2024autoregressive} to update memory units, a method widely used in recent vision trackers focused on temporal modeling. The visual memory feature $ M $ in MSM consists of a list of $ L_m $ memory units $ m $, denoted as $ M = \left\{ m_i \right\}_{i=1}^{L_m} $. Below, we will illustrate how the sliding windows memory storage method is implemented.

For a video sequence with $ T $ frames ($0 \leq t \leq T-1$), the memory units in $ M $ need to be initialized when processing the first frame (\ie, $ t=0 $). Specifically, after encoding the visual input information via a vision encoder, we obtain the feature $ f^0_{[C]} $ encoded from the [CLS] token. Considering that the [CLS] token can represent global visual features \cite{devlin2018bert}, we use $ f^0_{[C]} $ to initialize the $ L_m $ memory units.
During the time interval $ t \in [1, T-1] $, after tracking each search frame, we obtain the updated memory unit $ m^t $. We pop the memory unit with index 0 from $ M $ and append $ m^t $ to the end of $ M $. 

    



\begin{table}
\centering
\begin{tabular}{c|cccc}
        \toprule
          Model & Params & Speed & AUC & P\\
        \midrule
         JointNLT \cite{zhou2023joint} & 153M & 31FPS & 56.9 & 58.1\\
         MMTrack  \cite{zheng2023towards} & 177M & 37FPS & 58.6 & 59.4 \\
         MemVLT \cite{feng2025memvlt} & 175M & 32FPS & 63.3 & 67.4 \\
         ATCTrack-B  & 160M & 35FPS & 67.5 & 73.6 \\
         ATCTrack-L  & 340M & 30FPS & 68.6 & 75.0 \\
        \bottomrule
        \end{tabular}
\caption{Results of efficiency analysis.}
\label{table:efficiency_analysis}
\end{table}

\section{More Details on Model Implementation \label{app_ab_imp}}
Due to space constraints, only core model implementation details are provided in Sec. 4.1. Here, we supplement some additional details.
First, regarding the model structure, when performing context words calibration, we use two stacked modules consisting of Eq. (3) and Eq. (4). When executing visual memory representation, we use two stacked modules consisting of Equations Eq. (6) and Eq. (7). 
It is important to note that we only use the FFN in the visual memory representation part. Considering that the computational cost of FFN in Transformer modules is higher than that of Attention \cite{Transformer}, our module design helps reduce the overall parameters and computation of the model.

Additionally, for model training, we use the AdamW optimizer \cite{loshchilov2017decoupled} to optimize our model. The text encoder remains frozen, the learning rate is set to $10^{-5}$ for the vision encoder, $10^{-4}$   for the remaining unfrozen modules, and the weight decay is set to $10^{-4}$ . 
We train for a total of 150 epochs and reduce the learning rate by a factor of 10 after 120 epochs.
Finally, during the model inference stage, dynamic template updating follows the implementation of STARK \cite{STARK}. We set the update interval to 25 and the update confidence threshold to 0.8.

\section{Experimental Details of Ablation Studies \label{app_ab_1}}
In Sec. 4.3, we conduct detailed ablation analyses to investigate the properties of the various modules in ATCTrack. Due to space limitations, we do not fully elaborate on the specific implementation of the ablation experiments. In this section, we provide additional details.

\subsection{Ablation Study on important model components}
Tab. 2 presents the ablation study results of two core components in our approach: the textual and the visual target-context guidance modules. The specific implementations are as follows:

Tab. 2 (\#1) demonstrates the baseline results without our textual and visual object-context guidance modules. In this setup, textual features are processed as a whole entity, an approach widely adopted by recent trackers such as SNLT \cite{wang2023unified} and MMTrack \cite{zheng2023towards}. Specifically, we employ a transformer-based decoder to facilitate interaction between textual features $f_L$ and search features $f^t_X$:
\begin{equation}
    f^t_{R} = Trans_{Dec}(f^t_{X}, f_L),
    \label{equ:l_ad}
\end{equation}
where $Trans_{Dec}$ represents the standard transformer decoder layer~\cite{Transformer}, primarily consisting of attention operations and feed-forward networks. $f^t_{R}$ denotes the search features embedded with textual cues, which are subsequently fed into the prediction head to obtain final tracking results. To ensure fair comparison, we configure the transformer decoder with four layers, matching the parameter count with the visual and textual object-context guidance module.

Tab. 2 (\#2) shows the results using only the textual object-context guidance module. In this implementation, we omit the visual memory guidance process and directly feed the output features $f^{t}_{XL}$ from the textual target-context guidance module into the prediction head to obtain final results.

Tab. 2 (\#3) presents the results using only our visual object-context guidance module. In this implementation, we employ a transformer-based decoder to guide the search features with textual information, which is formulated as:
\begin{equation}
    f^t_{XL} = Trans_{Dec}(f^t_{X}, f_L),
    \label{equ:l_ad_1}
\end{equation}
For fair comparison, we implement a two-layer decoder architecture.

Tab. 2 (\#4) demonstrates the results of our complete ATCtrack model.

\begin{table*}[t]
    \centering
    \begin{tabular}{l|ccc|ccc|ccc|ccc}
    \toprule
     \multicolumn{1}{c|}{\multirow{2}{*}{Method}}
      & \multicolumn{3}{c|}{MGIT (Action)} & \multicolumn{3}{c|}{TNL2K} &\multicolumn{3}{c|}{LaSOT} & \multicolumn{3}{c}{LaSOT\(_{ext}\)} \\ \cline{2-13}
     & AUC & P${_{\text{Norm}}}$ & P & AUC & P${_{\text{Norm}}}$ & P & AUC & P${_{\text{Norm}}}$ & P & AUC & P${_{\text{Norm}}}$  &P  \\
      \bottomrule
      \multicolumn{5}{l}{\textit{Basic Variants}} \\
     \midrule
     Wang \protect\cite{wang2018describe} & - & - & - & - & - & - & 27.7 & - & 30.4 & - & -& - \\
      Feng \protect\cite{feng2019robust} & - & - & - & 25.0 & 34.0 & 27.0 & 50.0 & - & 56.0 & - & -& - \\ 
      Feng \protect\cite{feng2020real} & - & - & - & 25.0 & 33.0 & 27.0 & 35.0 & - & 35.0 & - & -& - \\ 
      GTI \protect\cite{GTI} & - & - & - & - & - & - & 47.8 & - & 47.6 & - & - & - \\
      TNL2K-II \protect\cite{TNL2K} & - & - & - & 42.0 & 50.0 & 42.0 & 51.3 & - & 55.4 & - & - & - \\
      SNLT \protect\cite{SNLT} & 3.6 & 22.6 & 0.4 & - & - & - & 54.0 & 63.6 & 57.4 & - & - & - \\ 
      VLT$_{\rm{TT}}$ \protect\cite{VLT} & 46.8 & 60.2 & 31.8 & 54.7 & 71.8 & 55.3 & 67.3 & 80.2 & 71.5 & 48.4 & 59.9 & 54.3 \\ 
      TransVLT \cite{zhao2023transformer} & - & - & - & 56.0 & 61.7 & - & 66.4 & - & 70.8 & - & - & - \\ 
      JointNLT \protect\cite{zhou2023joint} & 61.0 & 78.6 & 44.5 & 56.9 & 73.6 & 58.1 & 60.4 & 69.4 & 63.6 & - & - & - \\
      TransNLT \protect\cite{wang2023unified} & - & - & - & 57.0 & 75.0 & 57.0 & 60.0 & - & 63.0 & - & - & - \\
      DecoupleTNL \protect\cite{ma2023tracking} & - & - & - & {56.7} & {-} & {56.0} & 71.2 & {-} &{75.3} & -& - & - \\
      All-in-One \protect\cite{zhang2023all} & - & - & - & 55.3 & - & 57.2 & 71.7 & 82.4 & 78.5 & 54.5 & 63.5 & - \\ 
      MMTrack \protect\cite{zheng2023towards} & - & - & - & 58.6 & 75.2 & 59.4 & {70.0} & 82.3 & 75.7& 49.4 & 59.9 & 55.3 \\ 
      QueryNLT \protect\cite{shao2024queryvlt} & - & - & - & {56.9} & {73.6} & {58.1} & {59.9} & {69.6} &{63.5} & - & - & -\\ 
      TTCTrack \protect\cite{mao2024textual} & - & - & - & {58.1} & {-} & {-} & {67.6} & {-} &{-} & 48.8 & - & -\\ 
      OSDT \protect\cite{zhang2024one} & - & - & - & {\color{blue}{59.3}} & {\color{blue}{76.2}} & {\color{blue}{61.5}} & {64.3} & {73.4} &{68.6} & - & - & -\\ 
      OneTracker \protect\cite{hong2024onetracker} & - & - & - & {58.0} & {-} & {59.1} & {70.5} & {79.9} & 76.5 & - & - & -\\ 
      UVLTrack-B \protect\cite{ma2024unifying} & - & - & - & 62.7 & - & 65.4 & 69.4 & - & 74.9 &  49.2 & - & 55.8 \\ 
      CTVLT \protect\cite{feng2024enhancing} & 69.2 & - & 62.9 & 62.2 & - & 79.5 & 72.3 & - & 79.7 & - & - & - \\ 
      ChatTracker-B \protect\cite{sun2025chattracker} & - & - & - & 59.6 & 76.3 & 62.1 & 71.7 & 80.9 & 77.5 & - & - & - \\ 
      MemVLT \protect\cite{feng2025memvlt} & {\color{blue}69.4} & {\color{blue}81.3} & {\color{blue}63.7} & 63.3 & {\color{blue}80.9} & 67.4 & 72.9 & 85.7 & 80.5 & 52.1 & 63.3 & 59.8 \\
      SUTrack-B224 \protect\cite{chen2024sutrack} & - & - & - & 65.0 & - & 67.9 & 73.2 & 83.4 & 80.5 & {\color{blue}53.1} & {\color{blue}64.2} & {\color{blue}60.5} \\ 
      SUTrack-B384 \protect\cite{chen2024sutrack} & - & - & - & {\color{blue}65.6} & - & {\color{blue}69.3} & {\color{blue}74.4} & {\color{blue}83.9} & {\color{blue}81.9} & 52.9 & 63.6 & 60.1 \\ 
      \midrule[0.1pt]
      \rowcolor{gray!20}
      \textbf{ATCTrack-B} & {\color{red}73.7} &  {\color{red}84.5} & {\color{red}70.1} & {\color{red}67.5} & {\color{red}85.3} & {\color{red}73.6} & {\color{red}74.6} & {\color{red}87.0} &{\color{red}82.1} & {\color{red}54.6} & {\color{red}65.7} & {\color{red}62.8} \\ 
    \bottomrule
     \multicolumn{5}{l}{\textit{Performance-oriented Variants}} \\
     \midrule
     ChatTracker-L \protect\cite{sun2025chattracker} & - & - & - & 65.4 & {\color{blue}76.5} & 70.2 & 74.1 & 83.8 & 81.2 & - & - & - \\
     UVLTrack-L \protect\cite{ma2024unifying} & - & - & - & 64.8 & - & 68.8 & 71.3 & - & 78.3 &  51.2 & - & 59.0 \\ 
     SUTrack-L224 \protect\cite{chen2024sutrack} & - & - & - & 66.7 & - & 70.3 & 73.5 & 83.3 & 80.9 & {\color{blue}54.0} & {\color{blue}65.3} & {\color{blue}61.7} \\ 
     SUTrack-L384 \protect\cite{chen2024sutrack} & - & - & - & {\color{blue}67.9} & - & {\color{blue}72.1} & {\color{red}75.2} & {\color{blue}84.9} & {\color{red}83.2} & 53.6 & 64.2 & 60.5 \\ 
     \midrule[0.1pt]
      \rowcolor{gray!20}
      \textbf{ATCTrack-L} & {\color{red}74.0} &  {\color{red}86.5} & {\color{red}76.1} & {\color{red}68.6} & {\color{red}85.8} & {\color{red}75.0} & {\color{blue}74.7} & {\color{red}87.1} &{\color{blue}82.3} & {\color{red}55.4} & {\color{red}66.8} & {\color{red}64.0} \\
     \bottomrule
    \end{tabular} 
    \caption{Comparison with state-of-the-art vison-language trackers on four popular benchmarks: MGIT \protect\cite{hu2023multi}, TNL2K \protect\cite{TNL2K}, LaSOT \protect\cite{LaSOT}, and LaSOT\(_{ext}\) \cite{fan2021lasot}. 
    The best two results are highlighted in {\color{red}red} and {\color{blue}blue}, respectively.}
    \label{tab:app_vlts_sota}
\end{table*}

\begin{figure*}[ht]
\begin{center}
\centerline{\includegraphics[width=\textwidth]{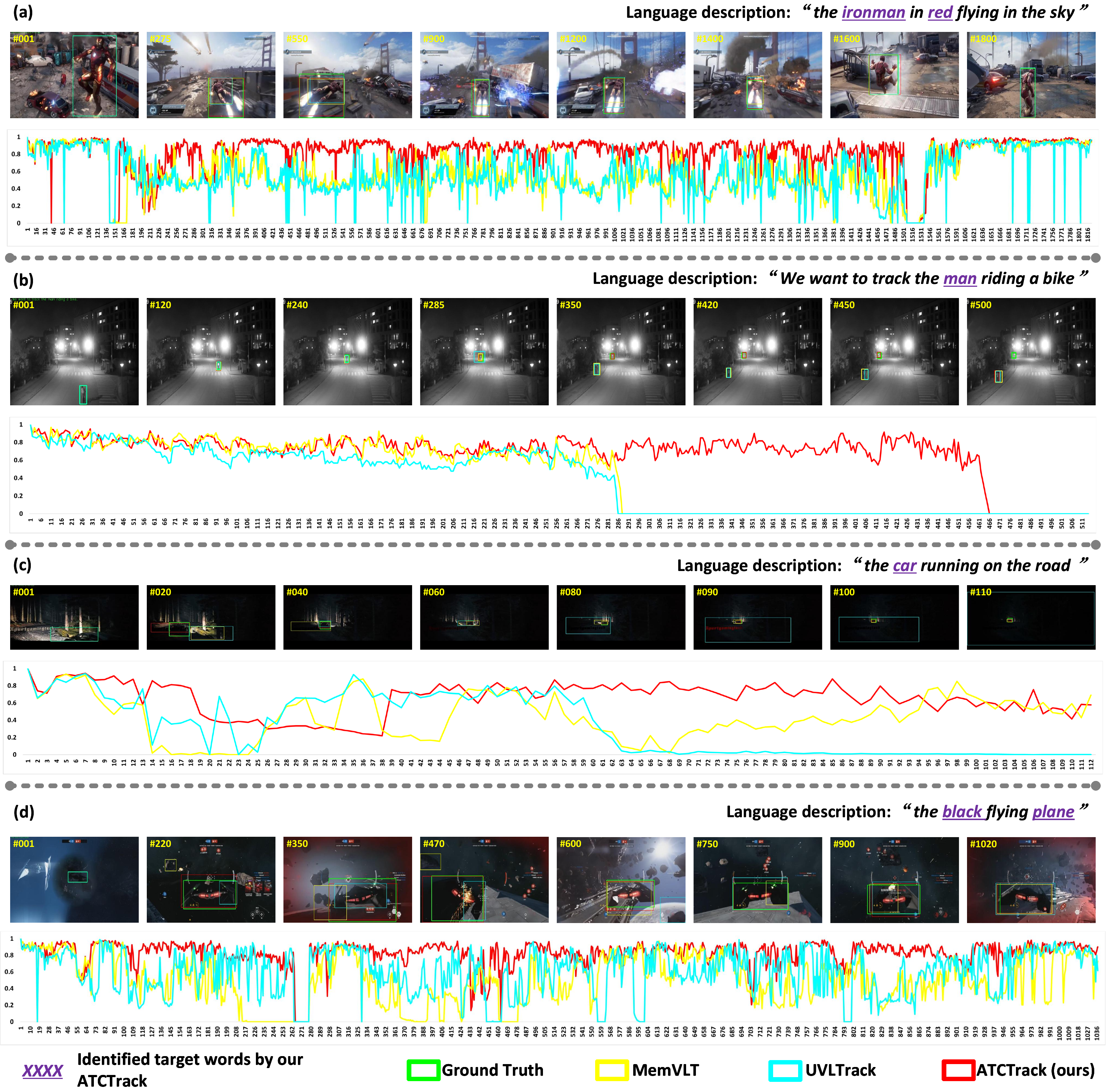}}
\caption{
Qualitative comparison results of our tracker with other two state-of-the-art vision-language trackers (\ie, MemVLT and UVLTrack) on four challenging cases. 
For each video case, we select representative frames to illustrate the predicted bounding boxes of each model and plot the curves of the IOU predictions across the entire video.
Better viewed in color with zoom-in.  
}
\label{fig_case_more}
\end{center}
\vskip -0.2in
\end{figure*}

\begin{table*}[]
    \centering
 \begin{tabular}{l|ccc|ccc|ccc}
    \toprule
     \multicolumn{1}{c|}{\multirow{2}{*}{Method}} 
     & \multicolumn{3}{c|}{TNL2K} & \multicolumn{3}{c|}{LaSOT} & \multicolumn{3}{c}{LaSOT$_{ext}$} \\ \cline{2-10}
     & AUC & P${_{\text{Norm}}}$ & P & AUC & P${_{\text{Norm}}}$ & P & AUC & P${_{\text{Norm}}}$ & P \\
    \bottomrule
      \multicolumn{5}{l}{\textit{Basic Variants}} \\
     \midrule
     SiamFC \cite{SiamFC} & - & - & - & 29.5 & 45.0 & 28.6 & 33.6 & 42.0 & 33.9 \\
       SiamRPN++ \cite{SiamRPNplusplus} & - & - & - & 41.3 & 48.2 & 41.2 & 49.6 & 56.9 & 49.1 \\
      SiamBAN \cite{SiamBAN} & - & - & - & 41.0 & 48.5 & 41.7 & 51.4 & 59.8 & 52.1 \\
      TransT \cite{TransT} & - & - & - & 64.9 & 73.8 & 69.0 & - & - & - \\
      Stark \cite{STARK} & - & - & - & 67.1 & 77.0 & - & - & - & -\\
      KeepTrack \cite{keeptrack} & - & -  & - & 67.1 & 77.2 & 70.2 & - & - & - \\
      Mixformer \cite{cui2022mixformer} & - & - & - & 69.2 & 78.7 & 74.7 & - & - & -  \\
      TransInMo \cite{guo2022learning} & 52.0 & {\color{blue}58.5} & 52.7 & 65.7 & 76.0 & 70.7 &  - & - & - \\
      OSTrack-256 \cite{OSTrack} & 54.3 & - & - & 69.1 & 78.7 & 75.2 & 47.4 & 57.3 & 53.3\\
      OSTrack-384 \cite{OSTrack} & 55.9 & - & - & 71.1 & 81.1 & 77.6 & 50.5 & 61.3 & 57.6\\
      AiATrack \cite{gao2022aiatrack} & - & - & - & 69.0 & 79.4 & 73.8 & 47.7 & 55.6 & 55.4\\
      SimTrack \cite{chen2022backbone} & - & - & - & 69.3 & 78.5 & - & - & - & - \\
      GRM \cite{gao2023generalized} & - & - & - & 69.9 & 79.3 & 75.8 & - & - & - \\
      SeqTrack-B256 \cite{chen2023seqtrack} & 54.9 & - & - & 69.9 & 79.7 & 76.3 & 49.5 & 60.8 & 56.3 \\
      SeqTrack-B384 \cite{chen2023seqtrack} & 56.4 & - & - & 71.5 & 81.1 & 77.8 & 50.5 & 61.6 & 57.5 \\
      ARTrack-256 \cite{wei2023autoregressive} & 57.5 & - & - & 70.4 & 79.5 & 76.6 & 46.4 & 56.5 & 52.3\\
      ARTrack-384 \cite{wei2023autoregressive} & 59.8 & - & - & 72.6 & 81.7 & 79.1 & 51.9 & 62.0 & 58.5\\
      OSTrack-Zoom \cite{kou2023zoomtrack} & 56.5 & - & 57.3 & 70.2 & - & 76.2 & 50.5 & - & 57.4 \\
      DropTrack \cite{wu2023dropmae} & 56.9 & - & 57.9 & 71.8 & 81.8 & 78.1 & 52.7 & 63.9 & 60.2\\
      ROMTrack-256 \cite{cai2023robust} & - & - & - & 69.3 & 78.8 & 75.6 & 48.9 & 59.3 & 55.0\\
      ROMTrack-384 \cite{cai2023robust} & - & - & - & 71.4 & 81.4 & 78.2 & 51.3 & 62.4 & 58.6\\
      F-BDMTrack-256 \cite{yang2023foreground} & 56.4 & - & 56.5 & 69.9 & 79.4 & 75.8 & 47.9 & 57.9 & 54.0\\
      F-BDMTrack-384 \cite{yang2023foreground} & 57.8 & - & 59.4 & 72.0 & 81.5 & 77.7 & 50.8 & 61.3 & 57.8\\
      EVPTrack-224 \cite{shi2024explicit} & 57.5 & - & 58.8 & 70.4 & 80.9 & 77.2 & 48.7 & 59.5 & 55.1\\
      EVPTrack-384 \cite{shi2024explicit} & 59.1 & - & 62.0 & 72.7 & 82.9 & 80.3 & {\color{blue}53.7} & {\color{blue}65.5} & {\color{blue}61.9}\\
      ODTrack-B \cite{zheng2024odtrack} & 60.9 & - & - & 73.2 & 83.2 & 80.6 & 52.4 & 63.9 & 60.1\\
      AQATrack-256 \cite{xie2024autoregressive} & 57.8 & - & 59.4 & 71.4 & 81.9 & 78.6 & 51.2 & 62.2 & 58.9\\
      AQATrack-384 \cite{xie2024autoregressive} & 59.3 & - & 62.3 & 72.7 & 82.9 & 80.2 & 52.7 & 64.2 & 60.8\\
      ARTrackV2-256 \cite{bai2023artrackv2} & - & - & - & 71.6 & 80.2 & 77.2 & 50.8 & 61.9 & 57.7 \\
      ARTrackV2-384 \cite{bai2023artrackv2} & - & - & - & 73.0 & 82.0 & 79.6 & 52.9 & 63.4 & 59.1\\
      HIPTrack \cite{cai2023learning} & - & - & - & 72.7 & 82.9 & 79.5 & 53.0 & 64.3 & 60.6\\
      OneTracker \cite{hong2024onetracker} & 58.0 & - & 59.1 & 70.5 & 79.9 & 76.5 & - & - & - \\
      LoRAT-B224 \cite{LoRAT} & 58.8 & - & 61.3& 71.7 & 80.9 & 77.3 & 50.3 & 61.6 & 57.1 \\
      LoRAT-B378 \cite{LoRAT} & 59.9 & - & 63.7& 72.9 & 81.9 & 79.1 & 53.1 & 64.8 & 60.6 \\
      SUTrack-B224 \protect\cite{chen2024sutrack} & 65.0 & - & 67.9 & 73.2 & 83.4 & 80.5 & 53.1 & 64.2 & 60.5 \\ 
      SUTrack-B384 \protect\cite{chen2024sutrack} & {\color{blue}65.6} & - & {\color{blue}69.3} & {\color{blue}74.4} & {\color{blue}83.9} & {\color{blue}81.9} & 52.9 & 63.6 & 60.1 \\ 
      \midrule[0.1pt]
      \rowcolor{gray!20}
      \textbf{ATCTrack-B}
        & {\color{red} $67.5$}
        & {\color{red} $85.3$}
        & {\color{red} $73.6$}
        & {\color{red} $74.6$}
        & {\color{red} $87.0$}
        & {\color{red} $82.1$}
        & {\color{red} $54.6$}
        & {\color{red} $65.7$}
        & {\color{red} $62.8$}
        \\
    \bottomrule
     \multicolumn{5}{l}{\textit{Performance-oriented Variants}} \\
     \midrule
     ODTrack-L \cite{zheng2024odtrack} & 61.7 & - & - & 74.0 & 84.2 & {\color{blue}82.3} & 53.9 & 65.4 & 61.7 \\
     LoRAT-L224 \cite{LoRAT} & 61.1 & - & 65.1& 74.2 & 83.6 & 80.9 & 52.8 & 64.7 & 60.0 \\
      LoRAT-L378 \cite{LoRAT} & 62.3 & - & 67.0 & {\color{blue}75.1} & 84.1 & 82.0 & {\color{red}56.6} & {\color{red}69.0} & {\color{red}65.1} \\
      SUTrack-L224 \protect\cite{chen2024sutrack} & 66.7 & - & 70.3 & 73.5& 83.3 & 80.9 & 54.0 & 65.3 & 61.7 \\ 
     SUTrack-L384 \protect\cite{chen2024sutrack} & {\color{blue}67.9} & - & {\color{blue}72.1} & {\color{red}75.2} & {\color{blue}84.9} & {\color{red}83.2} & 53.6 & 64.2 & 60.5 \\ 
     \midrule[0.1pt]
      \rowcolor{gray!20}
      \textbf{ACTrack-L} 
            & {\color{red} $68.6$}  & {\color{red} $85.8$} & {\color{red} $75.0$} & $74.7$ & {\color{red} $87.1$} & {\color{blue} $82.3$} & {\color{blue} $55.4$} & {\color{blue} $66.8$}  & {\color{blue} $64.0$}
            \\
     \bottomrule
    \end{tabular} 
    \caption{Comparison with state-of-the-art vision-only trackers on three popular benchmarks: TNL2K \protect\cite{TNL2K}, LaSOT \protect\cite{LaSOT}, and LaSOT\(_{ext}\) \cite{fan2021lasot}. 
    The best two results are highlighted in {\color{red}red} and {\color{blue}blue}, respectively.}
    \label{tab:app_results_sota}
\end{table*}

\subsection{Ablation Study on Textual Target-Context Modeling}
Tab. 3 shows different ways of utilizing textual cues, with the specific implementations for each setting as follows:

\paragraph{\textit{Naive method.}} This setting is consistent with that of Tab. 2 (\#1).

\paragraph{\textit{+ Target words awareness.}} This refers to the incorporation of target words awareness method based on the “\textit{naive method}" setting. 
Specifically, we concatenate the $f_{LT}$ with $f_{L}$ to obtain context features $f_{LC}$ for subsequent textual guidance.

\paragraph{\textit{+ Context words calibration.}} This refers to the incorporation of context words calibration operations based on the “\textit{+ target words awareness}” setting. This is the approach adopted by our ATCTrack.

\paragraph{\textit{- Dual-type textual guidance.}} This approach utilizes only the calibrated single-type text features $f_{L^{\prime}}$ for textual guidance, where $f_{LC} = f_{L^{\prime}}$.

\subsection{Ablation Study on Visual Target-Context Modeling}
Tab. 4 shows different ways of utilizing visual cues, with the specific implementations for each setting as follows:

\paragraph{\textit{Naive method.}} This setting is consistent with that of Tab. 2 (\#1).

\paragraph{\textit{+ ROI.}} This represents the augmentation of the “\textit{naive method}" by incorporating explicit visual memory features for tracking assistance. Specifically, we employ the Region of Interest (RoI) approach \cite{ren2015faster}, which is widely adopted in recent Visual-Language Trackers (VLTs) such as JointNLT \cite{zhou2023joint} and TrDiMP \cite{wang2021transformer}. We apply RoI processing to the search features $f^t_X$ using the predicted bounding box scaled by 1.5 to obtain localized search features $f^t_{X^{\prime}} \in \mathbb{R}^{36 \times D}$. Subsequently, the visual memory representation process is implemented through the following computations:
\begin{equation}  
     f_{[C]M^{\prime}}= Norm(f_{[C]M}+ \Phi_{CA}(f_{[C]M}, f^t_{X^{\prime}})),
\end{equation}  
\begin{equation}  
    f_{[C]M^{\prime\prime}}= Norm(f_{[C]M^{\prime}}+ FFN(f_{[C]M^{\prime}})).
\end{equation}

\paragraph{\textit{+ Search + crop mask.}} This setting involves using a local mask to construct the object-context indication map. 
Specifically, for the global object-context indication map \( h^t \), we retain only the values within the area corresponding to 1.5 times the predicted bbox, while setting the values in all other areas to zero, resulting in \( h^t_l \).
Then, the visual memory representation process is implemented through the following computations:

\begin{equation}  
     f_{[C]M^{\prime}}= Norm(f_{[C]M}+ \Phi_{CA}(f_{[C]M}, h^t_l\odot f^t_X )),
\end{equation}  
\begin{equation}  
    f_{[C]M^{\prime\prime}}= Norm(f_{[C]M^{\prime}}+ FFN(f_{[C]M^{\prime}})).
\end{equation}

\paragraph{\textit{+ Search + global mask.}} This setting involves using a global mask to construct the object-context indication map, which is used to obtain explicit visual memory features. This is the approach adopted by our ATCTrack.

\subsection{Ablation Study on the Contribution of different modules}
\paragraph{\textit{ w/o HiViT backbone.}} This setting refers to replacing the HiViT backbone \cite{zhang2022hivit,tian2024fast} with the ViT backbone typically used in conventional trackers \cite{OSTrack,cui2022mixformer}.

\paragraph{\textit{ w/o dynamic template.}} This setting refers to using only the original static template for visual input, without the sparse dynamic template \cite{STARK}.

\paragraph{\textit{ w/o Textual\(_{TC}\) \& Visual\(_{TC}\).}}
This setting is the same as setting in Tab. 2 (\#1), meaning that the visual and textual target-context guidance mechanism we designed is not utilized.

\paragraph{\textit{ w/o target words label.}}
This setting, with the model structure unchanged, refers to not using target words supervision signals, thus excluding $L_{\text{bce}}$ loss.

\section{Additional Experimental Results \label{add_exp}}
\subsection{Efficiency Analysis \label{add_fps}}

In \cref{table:efficiency_analysis}, we compare ATCTrack with the latest VLTs (\ie, JointNLT \cite{zhou2023joint}, MMTrack \cite{zheng2023towards}, and MemVLT \cite{feng2025memvlt}) in terms of efficiency (Params and Speed) and performance (AUC and P on TNL2K). For ATCTrack-B, the parameters and tracking speed are comparable to recent trackers, but it shows significant performance advantages, such as a 4.2\% improvement in AUC compared to MemVLT. 
For ATCTrack-L, the parameter scale is considerably larger than ATCTrack-B, which leads to a further performance improvement.


\subsection{Comparison with More Trackers}
In Tab. 1 of Sec. 4.2, due to space constraints, we compare ATCTrack with several recent high-performance vision-language trackers. 
As a supplement, \cref{tab:app_vlts_sota} presents the performance of a broader range of vision-language trackers. Additionally, in line with the prevailing paradigm of vision-language tracking models \cite{zhou2023joint, zheng2023towards, feng2025memvlt}, \cref{tab:app_results_sota} provides additional comparisons with vision-only trackers. The strong performance of our model among these trackers further demonstrates the effectiveness of our approach.

\section{More Qualitative Results \label{app_ab_visual_res}}
Due to space limitations, Fig. 4 only presents four cases for the qualitative comparison between our model and the latest SOTA models. In this section, we provide additional qualitative comparison results, as illustrated in Fig.~\ref{fig_case_more}.


\end{document}